\pdfoutput=1
\PassOptionsToPackage{section}{placeins}
\documentclass{mindlab}

\usepackage[utf8]{inputenc}
\usepackage{amsmath,amssymb,mathtools}
\usepackage{array,tabularx,longtable,threeparttable}
\usepackage{tikz}
\usetikzlibrary{arrows.meta,positioning,fit,backgrounds}
\usepackage{xurl}
\usepackage{enumitem}
\setlist{nosep,leftmargin=*}
\newcommand{\method}{MJev-Spatial}
\newcommand{\inlinecode}[1]{\nolinkurl{#1}}
\newcommand{\sourcepath}[1]{{\footnotesize\nolinkurl{#1}}}
\newcommand{\argmax}{\operatorname*{arg\,max}}
\newcommand{\R}{\mathbb{R}}
\DeclareRobustCommand{\authorcontactlist}{%
  \begin{tabular}[t]{@{}l@{\;}l@{}}
    Changhai Zhou: & \email{chzhou25@m.fudan.edu.cn} \\
    Andrew Chen: & \email{andrew@mindlab.ltd} \\
    Pony Ma: & \email{pony@mindlab.ltd}
  \end{tabular}}

\newlength{\mjevwordmarkcapheight}
\DeclareRobustCommand{\MJevSpatialTitle}{%
  \begingroup
  \settoheight{\mjevwordmarkcapheight}{M}%
  \raisebox{-0.32\mjevwordmarkcapheight}{%
    \includegraphics[height=1.39\mjevwordmarkcapheight]{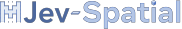}}%
  \endgroup}
\hypersetup{pdftitle={MJev-Spatial: System One Spatial Intelligence via Native Attention Matching}}

\title{\MJevSpatialTitle\hspace{0.10em}: System One Spatial Intelligence\\via Native Attention Matching}
\author[*]{Luozheng~Qin}
\author[*]{Yuhan~Zhu}
\author[*]{Changhai~Zhou}
\author{Qian~Qiao}
\author{Jun~Gao}
\author{Yuhua~Zhou}
\author{Harry~Zhang}
\author{Xiangyu~Zeng}
\author{Yuxuan~Liu}
\author{Yuxin~Lu}
\author{Kaijie~Chen}
\author{Andrew~Chen}
\author{Pony~Ma}
\contribution[*]{Equal contribution.}
\metadata[\faGithub\ Code]{\url{https://github.com/Fr0zenCrane/MJev-Spatial}}
\correspondence{\authorcontactlist}
\date{October 2026}
\abstract{

Spatial intelligence lets an agent perceive and act in the physical world, and many of its judgments must be made at interaction rates: which spatial relation holds, how far an object is, or where to place it.
Vision-language models (VLMs) produce these answers by free-form text generation, paying for autoregressive decoding and answer parsing even though the output space is compact: one of a few options, a scalar, or a 2D point.
In light of this, we introduce \textbf{MJev-Spatial}, a System~1 multimodal decision model that tackles all three types of spatial understanding tasks within a unified classification framework.
Specifically, MJev-Spatial treats each candidate as a class and scores it using the backbone's pretrained query and key projections, without adding parameters.
The candidates are the given options for choice tasks, or value bins and image regions refined through coarse-to-fine classification for numerical and point prediction, respectively.
At inference, each classification decision takes a single forward pass and generates no answer tokens.
To train on heterogeneous spatial data efficiently, we supervise all coarse-to-fine stages jointly with teacher forcing in one forward pass, and combine this with length-aware global batching and cross-GPU load balancing, raising training throughput by 30.8\% over the original batching strategy.
Built on Qwen3.5-4B and trained on 12.3M question-answer pairs converted from existing spatial datasets, MJev-Spatial achieves the highest overall score (65.64\%) among eight tested models, ranks first on seven of twelve benchmarks, and outperforms its unmodified autoregressive base model on all twelve, while running 2.5$\times$ faster than the fastest autoregressive baseline and 5.6$\times$ faster than the base model.

}

\input{frontmatter_layout}

\hypersetup{pdfauthor={Luozheng Qin, Yuhan Zhu, Changhai Zhou, Qian Qiao, Jun Gao, Yuhua Zhou, Harry Zhang, Xiangyu Zeng, Yuxuan Liu, Yuxin Lu, Kaijie Chen, Andrew Chen, Pony Ma}}
\begin{document}
\maketitle
\enlargethispage{\baselineskip}
\noindent\begin{minipage}{\textwidth}
\setlength{\parskip}{0pt}
\centering
\includegraphics[width=\textwidth]{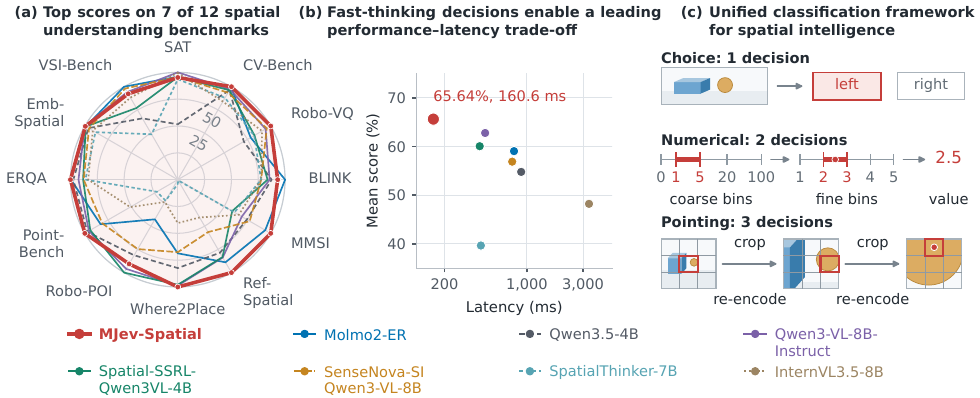}
\captionsetup{type=figure,font=footnotesize,skip=3pt}
\caption{
\captionlead{Overview of the proposed MJev-Spatial.}
(a) Top scores on 7 of 12 spatial understanding benchmarks. (b) The highest mean score and lowest mean latency among eight evaluated models. (c) Direct option selection and coarse-to-fine numerical and point prediction. MJev-Spatial unifies these three spatial task types within a classification framework, enabling fast-thinking decisions with high accuracy and low inference latency.}
\label{fig:teaser}
\end{minipage}

\par

\section{Introduction}
Spatial intelligence involves perceiving and reasoning about the geometry and spatial relationships of objects, which enables agents to navigate their surroundings and interact with the physical world~\citep{spatialvlm,spatialrgpt}.
Many of these interactions require timely spatial judgments.
Vision-language models (VLMs) support these capabilities by answering spatial questions grounded in visual observations.
Such questions may require selecting an option that describes a spatial relation~\citep{sat,cvbench,blink,mmsi}, estimating a numerical quantity such as distance~\citep{sensenova,cambrians,vst}, or identifying a point for object placement~\citep{refspatial,robopoint,robospatial,pointarena}.

To produce these spatial predictions, many existing VLMs retain their pretrained autoregressive text-generation capabilities~\citep{spatialvlm,spatialrgpt,molmo,robopoint,refspatial}.
Each option, numerical value, or image location is therefore represented as a token sequence that the model is trained to reproduce.
This couples learning the spatial answer with learning its textual form, including formatting and boilerplate when present, introducing supervision beyond the task-relevant prediction itself.
At inference, the model must generate the answer tokens sequentially, after which the resulting text must be parsed to recover the spatial prediction, adding decoding and parsing overhead.
This motivates a System~1 approach to spatial understanding: using pretrained visual and linguistic knowledge to make spatial judgments directly.
Jev provides a precedent for this approach by returning probabilities over specified candidates without generating answer text~\citep{jev}.
At the same time, we observe that all three common spatial output types can be represented as finite candidate sets: answer options are already finite, while numerical values and image locations can be discretized within a specified range and resolution.
Building on these insights, we introduce \textbf{MJev-Spatial}, a multimodal decision model specifically tailored to spatial understanding.

MJev-Spatial unifies the three output types by casting prediction as classification over finite candidate sets.
Each class represents a supplied answer option, a numerical bin, or an image region, depending on the task.
Selecting a class gives the chosen option, predicts which bin contains the target value, or predicts which region contains the target point.
To choose among these classes, MJev-Spatial encodes their descriptions together with the observation and question, then reuses the final attention layer's pretrained query and key projections to compute class scores.
A softmax over these scores gives the class probabilities, without additional prediction heads or trainable parameters.
At inference, each classification decision scores all its candidates in one forward pass.
Furthermore, to obtain fine-grained values and coordinates without evaluating a large number of classes at once, we subdivide the selected numerical bin or image region and classify again.
Numerical prediction uses up to two stages; single-point prediction uses three stages, re-encoding the selected image crop before each later decision to recover local visual detail.
The final selection is decoded to a numerical value or image coordinates.
Conditional tree expansion evaluates multiple region paths to extend the same trained point predictor to multiple points without additional training.

Numerical and point prediction can require multiple supervised stages, with new image crops introduced between point decisions.
Ground-truth answers determine the target classes and crops, allowing all stages to be jointly supervised in one teacher-forced causal forward pass.
We average the stage cross-entropy losses per question so that a multi-stage example does not receive greater weight than a one-stage choice.
Global length grouping and cross-GPU load balancing address the padding and workload imbalance caused by different sequence lengths and crop counts while preserving optimizer-batch membership.
An audited conversion pipeline supplies 12.3M QA from existing spatial annotations for full-parameter training of the retained language backbone.

Built on Qwen3.5-4B~\citep{qwen35}, MJev-Spatial achieves the highest overall score among eight evaluated models, leads on seven benchmarks, and outperforms its unmodified autoregressive base model on all twelve.
Across all twelve benchmarks, it achieves 2.5$\times$ and 5.6$\times$ speedups over the fastest autoregressive baseline and the base model, respectively.
Separate studies examine candidate scoring, visual refinement, numerical discretization, and training execution.
The readout and refinement studies assess how candidate scoring with pretrained projections and crop re-encoding contribute to prediction quality and computational cost; the numerical study shows that coarser length discretization can reduce prediction error despite its lower resolution.
Global length grouping and cross-GPU load balancing further improve training throughput by 30.8\% over the original batching strategy.

Our contributions are as follows:
\begin{enumerate}[leftmargin=*,label=(\arabic*)]
\item We propose a novel System~1 paradigm for spatial intelligence, unifying option selection, numerical estimation, and image pointing through direct classification over finite candidate sets.
\item Building on this paradigm, we develop MJev-Spatial and a unified training recipe, using pretrained attention projections for candidate scoring and joint supervision across spatial tasks and refinement stages. \item Across twelve spatial benchmarks, MJev-Spatial achieves the highest mean score among eight evaluated models and a 5.6$\times$ speedup over its autoregressive base model, Qwen3.5-4B.

\end{enumerate}

\section{Related Work}
\paragraph{System 1 decision model.}
Jev motivates predicting over declared answer spaces instead of generating unrestricted answer text~\citep{jev}.
Jev-Omni extends this interface to multimodal inputs~\citep{jevomni}, while KEV implements candidate scoring with learned query and key projections~\citep{kev}.
\citet{visualjev} and \citet{jevvisual} study direct decisions and shared-context execution; \citet{embodiedjev} applies decision interfaces to skill and action selection.
Concurrent work, VideoJev, uses a shared candidate scorer for temporal grounding and video question answering, with cached and joint inference for queries about the same video~\citep{zhu2026videojevmodelvideounderstanding}.
Visual Jev reads candidate-token probabilities from the existing language-model head and finds no consistent accuracy advantage from an added typed head in its matched control.
MJev-Spatial instead scores contextual candidate content with the backbone's final-layer query and key projections and trains all output types with the same candidate-classification objective.
Our matched readout experiments show competitive quality and lower training cost for this design, with task-dependent tradeoffs (\cref{sec:readout-results}).
The earlier Molmo2-ER-based MJev-Spatial prototype used LoRA and an added decision head~\citep{molmo2er,lora}.
The present model uses Qwen3.5-4B and trains all retained language parameters.
Backbone efficiency is also studied through pruning and quantization, joint low-rank and sparsity optimization, and adaptive execution~\citep{zhou2026buddy,zhou2026deputy,zhou2025qpruner, zhou2025largecompression}.

Class-aware adaptation provides a complementary route to efficient vision-language classification.
AWT enriches class descriptions and combines augmented views through entropy-based weighting and optimal transport~\citep{zhu2024awt}.
Self-TPT uses self-supervised prompt adaptation for new classes to avoid costly per-image optimization at inference~\citep{zhu2026efficient}.

\paragraph{Spatial understanding with VLMs.}
Broader multimodal research covers visual reasoning~\citep{qin2025unicot}, retrieval~\citep{qin2024bvrcc}, and multimodal generation~\citep{zhang2024musetalk,qin2026univigu,zhong2025anytalker}.
Video understanding further requires temporal localization~\citep{zhu2024dual,zhu2026timelens2} and efficient processing of long or streaming observations~\citep{li2026videochat3,zeng2026onestreamer}.
SpatialVLM and SpatialRGPT learn spatial relations and numerical answers from spatial supervision~\citep{spatialvlm,spatialrgpt}, while Molmo, RoboPoint, and RoboRefer support pointing through coordinate text~\citep{molmo,robopoint,refspatial}.
Alternative interfaces can ground predictions more directly.
MolmoPoint selects visual patches and refines locations through a sequence of specialized grounding tokens~\citep{molmopoint}.
MJev-Spatial uses one candidate-classification mechanism for options, numerical values, and points, with task-specific candidates supplying the output structure.
Its multi-point extension expands conditional classification paths rather than generating a variable-length coordinate string.

Representing continuous predictions with discrete classes is established in computer vision.
DORN formulates monocular depth estimation as ordinal regression over logarithmically spaced depth intervals~\citep{dorn}.
Our length representation also uses logarithmically spaced coarse intervals, followed by linear refinement, but its intervals are candidate descriptions scored by a shared language backbone.
Counts and angles retain their own ranges and discretization schemes.
Visual search methods such as V$^*$ refine image regions to recover details needed for answering~\citep{vstar}.
Video-o3 similarly acquires evidence iteratively in long videos, interleaving clue seeking with inspection of relevant segments~\citep{zeng2026video}.
In MJev-Spatial, region selection itself is supervised as classification, and re-encoded crops provide the evidence for subsequent decisions.

\section{Method}
\label{sec:method}
MJev-Spatial takes visual observations, a spatial question, and task-specific candidates as input.
Following Jev~\citep{jev}, it predicts by selecting candidates rather than generating text.
We first explain how one classification framework handles answer options, numerical values, and image locations, including coarse-to-fine refinement and multi-point prediction.
We then describe the shared model architecture and the training recipe.

\subsection{Unified Classification Framework for Spatial Intelligence}
\label{sec:transitions}
Given ordered images $I=(I_1,\ldots,I_m)$ and a spatial question $q$, the model predicts an option, a number with a declared unit, or a point in normalized image coordinates.
Each candidate is a possible answer for the current decision.
For a choice question, it is one of the supplied options; for a numerical or point question, it identifies an interval or region in which the target is predicted to lie.
This lets all three output types use the same candidate-classification rule.

\begin{figure}[!t]
\centering
\input{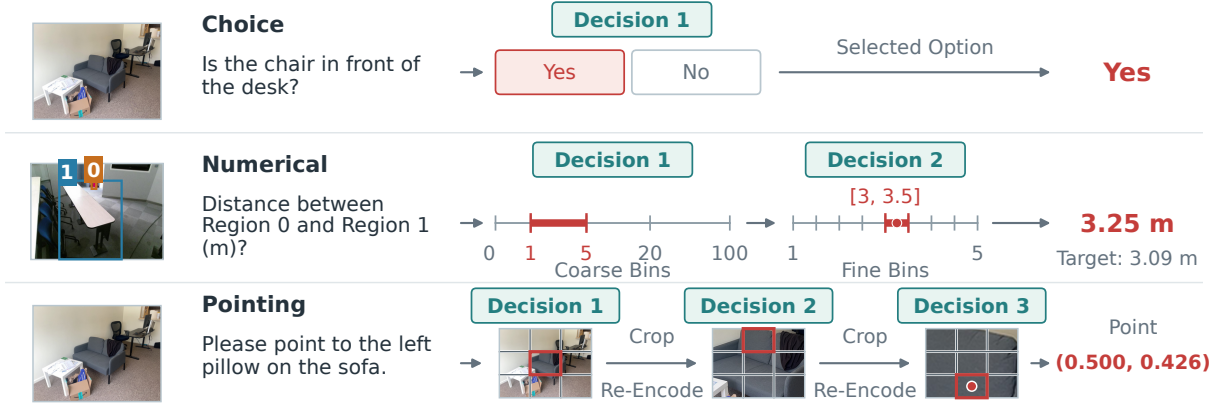}
\caption{\captionlead{Task formulation and coarse-to-fine inference.}
Every stage uses a shared candidate scorer.
Choice tasks select a supplied option. Numerical estimation selects a coarse interval and then a fine interval, returning its midpoint. Point prediction uses three $3\times3$ region selections, re-encoding crops between stages and returning the final cell center.}
\label{fig:task-inference}
\end{figure}

\begin{samepage}
At stage $t$, a deterministic task controller constructs candidates $\mathcal C_t=\{c_{t,1},\ldots,c_{t,K_t}\}$ from the question and preceding selections and visual refinements $h_{<t}$.
The shared scorer assigns a logit $z_{t,i}$ to each candidate, and prediction selects the highest-scoring class:
\begin{equation}
 p_\theta(i\mid I,q,h_{<t},\mathcal C_t)
 =\frac{\exp z_{t,i}}{\sum_{j=1}^{K_t}\exp z_{t,j}},
 \qquad \hat a_t=\argmax_i p_\theta(i\mid I,q,h_{<t},\mathcal C_t),
 \label{eq:choice}
\end{equation}
\par
\end{samepage}
Only the current legal candidates enter the softmax; padded slots receive $-\infty$.
A choice question ends after this decision.
For numerical and point prediction, the selected class determines the next candidate set and any new image crop.
The final class is decoded to an exact value or interval midpoint for numerical tasks, or a region center for pointing.
\Cref{fig:task-inference} illustrates these prediction paths, and \cref{tab:contracts} summarizes the output contracts.

Each classification decision scores all current candidates in one forward pass, with no answer-token generation or text parsing.
A complete prediction can require several such decisions: numerical tasks use up to two stages and single-point prediction uses three.
Multi-point prediction evaluates additional branches of the same spatial hierarchy.
The common scorer is unchanged across these tasks; their candidates and refinement rules supply the output structure.

\begin{table}[htbp]
\centering\small
\caption{Output types and their decision structure.
Numerical tasks share the scalar interface but keep separate codebooks.
Ordinary pointing produces one point; the probability tree is an evaluation extension for multi-point counting.}
\label{tab:contracts}
\begin{tabularx}{\textwidth}{@{}l l l X@{}}
\toprule
\rowcolor{mindlabtablehead}
\mindlabtablehead{Output} & \mindlabtablehead{Candidates per stage} & \mindlabtablehead{Stages} & \mindlabtablehead{Decoded answer} \\\midrule
Choice & Supplied option contents & 1 & The selected option, mapped back through the recorded option permutation.\\
Length (m) & 32 roots; 4 or 8 fine cells & 1--2 & Zero, overflow, or fine-cell midpoint.\\
Count & 32 coarse groups; 32 integers & 2 & $32a_1+a_2$, covering 0--1023 exactly.\\
Angle (deg) & 14 roots; 32 fine cells & 1--2 & Zero, overflow, or degree interval midpoint.\\
Point & 9 spatial cells at each level & 3 & Center of the final cell in original image coordinates.\\
\bottomrule
\end{tabularx}
\end{table}

\paragraph{Numerical prediction.}
\label{sec:numeric}
Numerical prediction must cover a wide range of values while retaining useful precision.
Enumerating every fine interval at once would make the candidate list large.
We instead construct candidates in two stages: the first selects a coarse interval and the second selects a fine cell within it.
Both stages use the scorer in \cref{sec:readout}; exact-zero and overflow classes terminate after the first decision.
The quantity specified by the question determines the unit and codebook.

\emph{Length codebook.}
Lengths cover $(0,U]$ with $U=16384$\,m through 30 positive coarse intervals, plus exact-zero and overflow classes.
With $s=0.01$\,m, the interval edges are
\begin{equation}
 e_j=s\left[\exp\left(\frac{j}{30}\log(1+U/s)\right)-1\right],
 \qquad j=0,\ldots,30.
 \label{eq:edges}
\end{equation}
This log1p spacing lets short dimensions and long distances share one meter-based codebook.
For a positive root, the fine stage splits $[e_j,e_{j+1}]$ linearly into $F_j$ cells, where $F_j$ is the smallest value in $\{4,8,16,32\}$ satisfying
\begin{equation}
 \frac{e_{j+1}-e_j}{2F_j}\leq
 \max(0.01\,\mathrm{m},0.05e_j).
 \label{eq:tolerance}
\end{equation}
This gives four cells in each of the first six intervals and eight in each of the remaining 24, 216 positive leaf cells in total.
Fine index $k$ decodes to the cell midpoint
\begin{equation}
 \hat y=e_j+\frac{k+1/2}{F_j}(e_{j+1}-e_j).
 \label{eq:scalar}
\end{equation}
Because the tolerance does not decrease within an interval, choosing the correct cell bounds the quantization error by $\max(1\,\mathrm{cm},5\%\,y)$; this bound concerns the codebook, not the model's prediction error.
Exact zero terminates immediately, and overflow terminates with representative $U$ on an explicit overflow path.
Candidate descriptions and decoding both read the codebook stored with the checkpoint, so training labels, interval text, and inference values cannot drift apart.

\emph{Decoding.}
We take the argmax at both stages and output the midpoint of the selected fine cell by \cref{eq:scalar}.
A conditional expectation over the fine cells of the selected root reads the same logits but gives higher error at the 72K scale (\cref{sec:length-results}), so we do not use it.

\emph{Counts and angles.}
Counts are discrete: $n\in[0,1023]$ is encoded as $(\lfloor n/32\rfloor,n\bmod32)$ and reconstructed exactly when both choices are correct.
Angles use twelve 30-degree positive intervals over $(0,360]$, 32 linear cells per interval, and zero and overflow roots, with the angle frame and direction declared by the question.
The adapter rejects unsupported quantities and ambiguous units, so meters, degrees, and counts never share a numerical interpretation.
Scalar counting is also kept separate from locating a set of points.

\paragraph{Point prediction.}
\label{sec:pointing}
Point prediction also requires fine resolution, but selecting a smaller region creates an additional need for local visual detail.
We therefore combine three nine-way region decisions with crop re-encoding: after each of the first two selections, a new crop provides the visual evidence for the next decision.
Let $B_t=[x_t,y_t,x_t+w_t,y_t+h_t]$ be the current rectangle in normalized original-image coordinates, with $B_0=[0,0,1,1]$.
For grid column $c\in\{0,1,2\}$ and row $r\in\{0,1,2\}$,
\begin{equation}
 B_{t+1}=\left[x_t+\frac{cw_t}{3},\ y_t+\frac{rh_t}{3},\;
 x_t+\frac{(c+1)w_t}{3},\ y_t+\frac{(r+1)h_t}{3}\right].
 \label{eq:region}
\end{equation}
The candidates follow a fixed order: top-center, top-left, top-right, center, middle-left, middle-right, bottom-center, bottom-left, bottom-right.
After each of the first two decisions, the selected rectangle is cropped from the original pixels, encoded by the same frozen visual pipeline, and appended with its decision history.
Cropping from the original rather than from the previous crop avoids compounding resampling.
The original view stays in context for global cues, while the crop spends visual tokens on the selected region.
Candidate text describes the current region, but coordinates always refer to the original image.

\begin{samepage}
After three decisions, the output is the center of $B_3$, which gives an effective $27\times27$ grid with a correct-cell center error of at most $1/54$ of the image width or height per axis.
An error at an earlier decision can exclude the correct region, so this geometric bound does not guarantee prediction accuracy.
Crop refinement costs two additional image encodings per point, and its benefit is tested against matched alternatives in \cref{sec:visual-results}.
\par
\end{samepage}

\paragraph{Extending to multiple points prediction.}
A single greedy path returns one location.
To recover several locations with the same trained model, we expand multiple paths and assign each final cell the product of its conditional probabilities along the path.
Together, these probabilities form a spatial map in which separated high-probability regions can yield multiple predictions.
We retain cells above a fixed threshold relative to the map's peak, group connected cells, and return the center of the highest-probability cell in each group.
The number of output points is determined by the resulting groups.
This requires no additional training or learned stop decision.
Full expansion visits 91 decision nodes and yields 729 cells; the evaluation decoder prunes low-probability root branches to reduce this cost.
Traversal and component extraction use model predictions only, with no target masks or true point counts.
\Cref{app:point-decoding} gives the fixed decoding rules, and \cref{sec:tree-results} evaluates their quality and computational tradeoffs.

\begin{figure}[!t]
\centering
\input{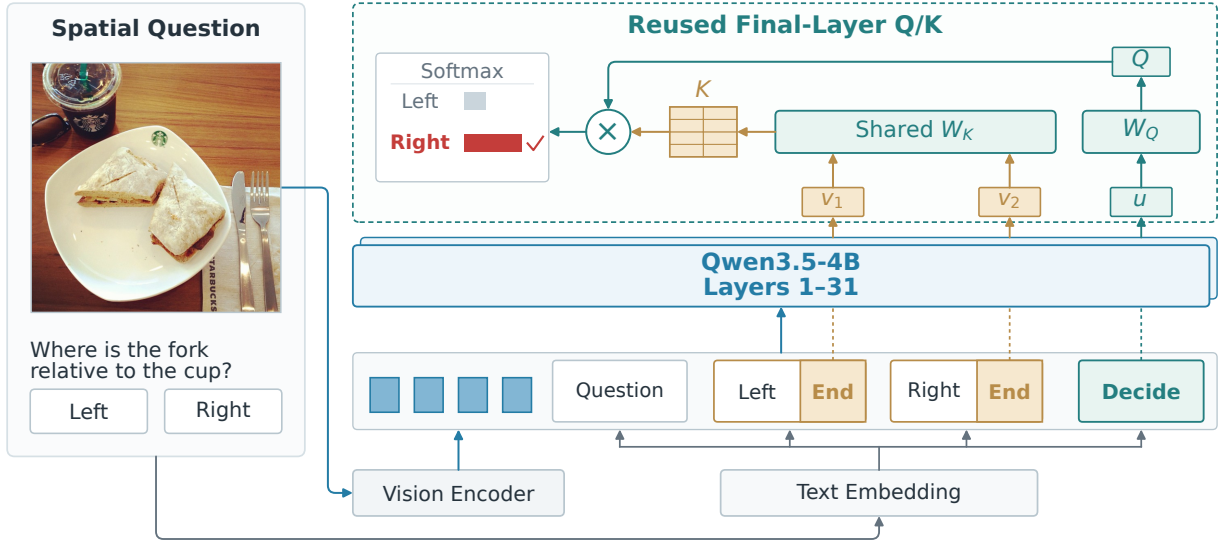}
\caption{\captionlead{Architecture of MJev-Spatial.}
Each candidate is followed by the same End marker. After contextualization by the language backbone, the state at each End position is passed through the same pretrained $W_K$ to form a separate candidate key; the final Decide state supplies the query through $W_Q$.
Their similarities determine the selected candidate, with no added prediction head or new parameters.
Native normalization, rotary transformations, and head averaging are retained but omitted from the diagram. Probability bars are schematic.}
\label{fig:architecture}
\end{figure}

\subsection{Model Architecture}
\label{sec:readout}
\label{sec:serialization}
The framework in \cref{sec:transitions} defines the candidate classes and refinement rules; a model must still determine which class is supported by the visual observation and question.
Following Jev~\citep{jev}, we adapt a pretrained VLM to make these decisions.
Its contextual representations draw on visual and linguistic knowledge acquired during large-scale pretraining, providing a basis for interpreting the scene and evaluating possible answers.
We build on this capacity by formulating classification as matching a decision representation with candidate representations.
Our design reuses both pretrained features and matching projections, giving the classifier a pretrained initialization without an additional prediction head.
\Cref{fig:architecture} illustrates the candidate sequence and the representations used for scoring.

Each candidate is expressed as text: choice tasks retain the supplied option content, numerical tasks specify values or intervals with their units, and pointing tasks name cells relative to the current image region, such as ``top-left'' or ``center''.
These descriptions provide the candidate content that the VLM contextualizes with the observation and question.

We instantiate MJev-Spatial with Qwen3.5-4B~\citep{qwen35}.
Its vision encoder and merger convert the images into visual tokens, which are encoded together with the question, preceding refinement history, and current candidate descriptions in a causal sequence.
Each candidate description ends with an option-end marker, and the complete candidate list ends with a decision marker.
After the first 31 language layers, the hidden state at each option-end marker represents that candidate in the context of the observation, question, and preceding candidates.
The hidden state at the decision marker incorporates the full candidate set and serves as the decision representation.
The markers reuse existing vocabulary entries, so no new embeddings are introduced.
\Cref{app:serialization} gives the exact sequence layout.

FreeRet finds that the final MLP can shift representations toward lexical prediction at the expense of semantic matching~\citep{zhu2025freeret}.
Motivated by this observation, we compute candidate scores before this transformation using the final full-attention layer's pretrained query and key projections.
The decision representation supplies the query, and each candidate representation supplies a key.
We retain the layer's pretrained input normalization, head normalization, and rotary position transformations.
Let the resulting vectors at stage $t$ be $Q_{t,h}$ and $K_{t,i,g(h)}$.
Candidate $i$ receives the mean scaled inner product across the query heads:
\begin{equation}
 z_{t,i}=\frac{1}{H}\sum_{h=0}^{H-1}
 \frac{Q_{t,h}^{\top}K_{t,i,g(h)}}{\sqrt d},
 \qquad H=16,\quad d=256,\quad g(h)=\lfloor h/4\rfloor.
 \label{eq:score-summary}
\end{equation}
Each group of four query heads shares one of the four key heads.
These scores are logits; a single softmax over the current candidates gives the class probabilities in \cref{eq:choice}.
The candidate descriptions change with the task and refinement stage, while the backbone and scoring rule remain shared.
\Cref{eq:native} specifies the projection and position operations in full.

All candidates for a decision are evaluated in the same forward pass.
We gather only the decision and option-end states before projection, and remove the unused value, gate, output, and feed-forward paths of the final layer.
All retained language parameters are trainable; the vision encoder and merger remain frozen, as described in the training recipe below.
Because candidate representations are causally ordered, training deterministically shuffles options, while evaluation follows the benchmark's specified ordering protocol.

\subsection{Training Recipe}
\label{sec:data}
We convert existing spatial annotations into 12.3M training QA and supervise all refinement stages of each question in one language-backbone forward pass.
Length grouping and cross-GPU load balancing reduce padding and workload imbalance.

\paragraph{Training data.}
Our training corpus contains 12.3M question--answer pairs converted from RefSpatial~\citep{refspatial}, SAT~\citep{sat}, SenseNova-SI~\citep{sensenova}, VSI-590K-image~\citep{cambrians}, and VST-P~\citep{vst}.
The training data cover multiple-choice spatial question answering (7.1M QA), point localization (3.5M QA), length estimation (1.6M QA), angle estimation (120K QA), and counting (38K QA).
These totals count original QA before expansion into decision stages.
See \cref{tab:data} in Appendix~\ref{app:data-details} for the source-by-task breakdown, with conversion and validation details provided in the same appendix.

\paragraph{Joint stage supervision.}
\label{sec:method-training}
Every supervised decision uses categorical cross-entropy over its current candidate set.
A choice question has one target class, whereas numerical and point questions can contain a sequence of target classes; point refinement also introduces image crops between stages.
The annotated answer determines the complete target path and any corresponding crops before training.
We serialize the stages in causal order, with each target choice and any subsequent crop placed after the corresponding decision position.
Earlier decisions therefore cannot attend to later choices or crops, while teacher forcing allows all stage losses to be evaluated in a single language-backbone forward pass.
At inference, the next candidates and crop depend on the model's current selection, so the stages are evaluated sequentially.
For a QA $q$ with $S_q$ supervised stages, the loss is
\begin{equation}
 \mathcal L=\frac1{N_{\rm actual}}\sum_q\frac1{S_q}
   \sum_{t=1}^{S_q}-\log p_\theta(a^*_{q,t}\mid
             I_q,q,h^*_{q,<t},\mathcal C_{q,t}).
 \label{eq:loss}
\end{equation}
Here $N_{\rm actual}$ is the number of real QA in the optimizer batch; any tail placeholders receive zero weight.
Averaging over stages gives each original QA equal total weight, so a three-stage point counts no more than a one-stage choice.

\paragraph{Efficient batching.}
Joint supervision removes the need for a separate language-backbone training pass at each refinement stage, but questions still require different amounts of computation.
In particular, point examples add two image crops, while choice and numerical examples have different candidate lengths and stage counts.
Padding dissimilar sequences wastes work, and assigning unequal work to GPUs delays synchronization.
We sort the expanded sequences within each global optimizer batch by length to form similar-length microbatches.
A longest-work-first assignment~\citep{lpt} then distributes the microbatches across GPUs, using a cost that accounts for padded sequence length and visual-token work.
This combines length grouping with cross-GPU load balancing while preserving the questions in each optimizer batch and their weights in \cref{eq:loss}.
Every QA remains an independent causal row: packing different questions together would additionally require isolation of Qwen's Gated DeltaNet states~\citep{gateddeltanet}.
The assignment rule and distributed normalization are specified in \cref{app:training-execution}.

Given the scale of our expanded training corpus, we adopt full-parameter fine-tuning of the language backbone, although parameter-efficient fine-tuning (PEFT) is also applicable~\citep{zhou2025rankadaptor,zhou2026lara,zhou2025bslora,zhou2025balancing,zhou2026autoqra}.
The final configuration uses eight GPUs, a per-device microbatch size of eight, and eight gradient accumulation steps, yielding a global batch size of 512.
We use AdamW~\citep{adamw} with learning rates of $5\times10^{-6}$ for the backbone and $2\times10^{-4}$ for the readout.
Language computation uses BF16, while trainable weights, gradients, optimizer states, and candidate scoring remain in FP32.
The vision encoder and merger are frozen.
\Cref{sec:formal} provides the complete configuration, and \cref{sec:cookbook} evaluates the scheduling improvements under controlled conditions.

\begin{table}[t]
\centering
\footnotesize
\setlength{\tabcolsep}{1.8pt}
\renewcommand{\arraystretch}{1.16}
\renewcommand{\tabularxcolumn}[1]{m{#1}}
\newcommand{\mainboardheader}[2][\centering]{%
  \parbox[c][2.2\baselineskip][c]{\linewidth}{#1\mindlabtablehead{#2}}}
\begin{threeparttable}
\caption{
Performance comparison of MJev-Spatial with seven autoregressive VLMs on 12 spatial understanding \mbox{benchmarks}.
Best and second-best scores are bold and underlined, respectively.
}
\label{tab:release-mainboard}
\label{tab:final-formal-results}
\begin{tabularx}{\textwidth}{@{}>{\raggedright\arraybackslash}m{.23\textwidth}*{13}{>{\raggedleft\arraybackslash}X}@{}}
\toprule
\rowcolor{mindlabtablehead}
\mainboardheader[\raggedright]{Model} & \mainboardheader{SAT} & \mainboardheader{CV-\newline Bench} & \mainboardheader{Robo-\newline VQ} & \mainboardheader{BLINK} & \mainboardheader{MMSI} & \mainboardheader{Ref\newline Spatial} & \mainboardheader{Where\newline 2Place} & \mainboardheader{Robo-\newline POI} & \mainboardheader{Point-\newline Bench} & \mainboardheader{ERQA} & \mainboardheader{Emb\newline Spatial} & \mainboardheader{VSI\newline full} & \mainboardheader{Overall} \\\midrule
\rowcolor{mindlabtablehighlight}
{\bfseries MJev-Spatial} & 79.33 & {\bfseries 88.82} & {\bfseries 79.06} & \underline{67.17} & {\bfseries 45.60} & {\bfseries 58.50} & {\bfseries 67.96} & 55.74 & {\bfseries 64.38} & {\bfseries 46.75} & \underline{78.46} & 55.93 & {\bfseries 65.64} \\
\addlinespace[1.5pt]
Molmo2-ER & 80.00 & \underline{88.31} & 62.01 & {\bfseries 72.17} & \underline{42.90} & \underline{52.00} & 46.64 & 26.23 & 53.47 & \underline{46.50} & 77.99 & {\bfseries 60.55} & 59.06 \\
\addlinespace[1.5pt]
Qwen3.5-4B & 43.00 & 85.56 & 56.17 & 63.11 & 33.90 & 46.00 & 55.98 & 50.00 & \underline{62.65} & 45.50 & 75.82 & 39.93 & 54.80 \\
\addlinespace[1.5pt]
\mbox{Qwen3-VL-8B-Instruct} & {\bfseries 83.33} & 86.37 & 74.72 & 61.76 & 31.40 & 49.00 & \underline{67.07} & \underline{59.02} & 61.54 & 43.25 & {\bfseries 78.90} & 57.07 & \underline{62.79} \\
\addlinespace[1.5pt]
\mbox{Spatial-SSRL-Qwen3VL-4B} & 79.67 & 85.52 & 65.26 & 60.39 & 26.70 & 48.75 & 66.54 & {\bfseries 61.48} & 60.60 & 41.25 & 78.10 & 46.65 & 60.08 \\
\addlinespace[1.5pt]
\mbox{SenseNova-SI Qwen3-VL-8B} & 79.67 & 83.15 & \underline{75.68} & 56.08 & 35.30 & 33.00 & 45.92 & 45.90 & 52.34 & 41.00 & 75.33 & \underline{59.38} & 56.90 \\
\addlinespace[1.5pt]
SpatialThinker-7B & 78.00 & 78.17 & 64.94 & 56.02 & 27.80 & 0.50 & 2.66 & 13.93 & 14.46 & 40.25 & 69.89 & 29.56 & 39.68 \\
\addlinespace[1.5pt]
InternVL3.5-8B & \underline{83.00} & 74.66 & 71.67 & 56.08 & 29.90 & 23.80 & 27.48 & 14.75 & 32.93 & 39.00 & 72.14 & 53.46 & 48.24 \\
\bottomrule
\end{tabularx}
\end{threeparttable}
\end{table}

\section{Experiments}
\label{sec:experiments}
We first compare the final MJev-Spatial model with seven autoregressive VLMs on spatial prediction quality and inference latency.
Controlled pilot experiments then compare candidate scorers, crop re-encoding and feature reuse, numerical discretization, and multi-point decoding.
Finally, we measure how length grouping and cross-GPU load balancing affect training throughput.

\subsection{Experimental Setup}
MJev-Spatial is compared with seven generative baselines: Qwen3.5-4B~\citep{qwen35}, Qwen3-VL-8B-Instruct~\citep{qwen3vl}, InternVL3.5-8B~\citep{internvl35}, Molmo2-ER~\citep{molmo2er}, Spatial-SSRL~\citep{spatialssrl}, SenseNova-SI~\citep{sensenova}, and SpatialThinker~\citep{spatialthinker}.
Our evaluation covers the popular SAT~\citep{sat}, CV-Bench~\citep{cvbench}, BLINK~\citep{blink}, MMSI-Bench~\citep{mmsi}, RefSpatial-Bench~\citep{refspatial}, Where2Place~\citep{robopoint}, Point-Bench~\citep{pointarena}, RoboSpatial VQA and pointing~\citep{robospatial}, ERQA~\citep{erqa}, EmbSpatial~\citep{embspatial}, and VSI-Bench~\citep{vsibench}.

For model-design ablations, we use a fixed 72K-QA mixture of 25K SAT choice questions, 22K VST-P length questions, and 25K RefSpatial point questions to reduce training cost while providing sufficient supervision for meaningful comparisons. 
Each model is trained for 2.2K updates using the same optimizer, learning-rate schedule, and image-token budgets.
VST-300 serves as the development set for length estimation.

\Needspace{23\baselineskip}
\subsection{Main Result}
\label{sec:release-results}

Our main evaluation tests the central premise of \method{}: a System~1 decision model can make accurate spatial judgments with low inference latency.
We therefore assess prediction quality and latency jointly, comparing \method{} with seven autoregressive VLMs.

\method{} achieves the highest average score of 65.64\%, ahead of 62.79\% for Qwen3-VL-8B-Instruct, and ranks first on seven of the twelve benchmarks (\cref{tab:release-mainboard}).
It also outperforms its autoregressive base model on all twelve benchmarks, including gains of 22.89 percentage points on RoboSpatial VQA and 12.50 points on RefSpatial-Bench.
These improvements span both spatial question answering and point localization.
For length estimation, the final model reaches an MAE of 0.103\,m on the VST-300 development set.
On an NVIDIA B200 GPU, \method{} also achieves the lowest mean query latency, 160.6\,ms, yielding a 2.5$\times$ speedup over the fastest autoregressive baseline and a 5.6$\times$ speedup over its base model (\cref{tab:final-inference-efficiency}).
The combination of the highest average score and lowest mean latency gives \method{} the strongest performance--latency trade-off among the compared models (\cref{fig:teaser}).
These results support our System~1 approach to spatial understanding: adapting pretrained multimodal knowledge into direct classification decisions that deliver both strong prediction quality and fast responses.

\subsection{Ablation Studies}
\label{sec:design-ablations}
We next examine the design choices that make direct classification effective across spatial tasks.
The studies follow the prediction process: we first assess the shared candidate-scoring mechanism, then examine how numerical discretization and visual refinement provide the precision needed for value and point prediction.
We further study how decoding adapts these predictions to different output requirements, including multiple points.
This progression connects the common classifier to the task-specific decisions behind the final model.

\paragraph{Pretrained candidate scoring.}
\label{sec:readout-results}
We compare native Q/K scoring with a typical linear classifier and a KEV~\citep{kev} pointer using newly initialized projections to assess whether adding a prediction head improves over reusing the pretrained projections (\cref{tab:aligned}).
The results show no consistent winner across tasks: native Q/K achieves the best RefSpatial accuracy, 51.00\%, and the lowest length-estimation MAE, 0.3341\,m, while the position classifier performs better on Where2Place and RoboSpatial pointing.
We therefore favor native Q/K for its minimal changes to the pretrained architecture: it reuses the existing projections without adding a prediction head or new trainable parameters.
It also has the fewest trainable parameters and the lowest measured training cost among the three variants.

\begin{table}[htbp]
\centering\small
\setlength{\tabcolsep}{5pt}
\caption{Comparison results on pretrained candidate scoring. ``Linear'' and ``KEV'' are abbreviation for linear classifier and KEV pointer, respectively.}
\label{tab:aligned}
\begin{tabular}{lrrr}
\toprule
\rowcolor{mindlabtablehead}
\mindlabtablehead{Metric} & \mindlabtablehead{Linear} & \mindlabtablehead{KEV} & \mindlabtablehead{Native Q/K} \\\midrule
SAT real & 78.33 & 76.33 & 78.00 \\
CV-Bench & 83.80 & 85.31 & 85.38 \\
RefSpatial & 49.00 & 45.50 & 51.00 \\
Where2Place & 67.99 & 61.02 & 59.98 \\
RoboSpatial pointing & 65.57 & 63.11 & 62.30 \\
RoboSpatial VQA & 60.34 & 68.54 & 68.06 \\
VST-300 MAE (m) & 0.3668 & 0.3614 & 0.3341 \\
\bottomrule
\end{tabular}
\end{table}

\Needspace{10\baselineskip}
\paragraph{Numerical discretization.}
\label{sec:length-results}
For numerical estimation, we examine whether finer candidate intervals improve prediction accuracy.
Keeping the same coarse intervals, we reduce the positive leaf cells from 960 to 216 by replacing 32 fine bins per interval with the 4/8-bin design.
Although the coarser codebook has higher quantization error, it reduces VST-300 dev MAE from 0.3277 to 0.3202\,m (\cref{tab:numerical-candidates}).
The result shows that finer representational precision does not necessarily translate into better learned predictions.
Using the same 216-cell model, replacing argmax decoding with a conditional expectation increases MAE to 0.3300\,m (\cref{tab:numerical-decoding}).
We therefore retain the compact codebook and argmax decoding, which give the lowest dev MAE in this comparison.

\begin{table}[H]
\centering
\caption{\textbf{Numerical prediction ablations.}
(a) Fine-stage candidate count with fixed coarse intervals and argmax decoding.
(b) Fine-stage decoding using the same 216-cell model, with coarse-stage argmax.
Quantization error uses training length labels; prediction metrics use the VST-300 dev set.
Shading marks the adopted settings.}
\label{tab:numerical-ablations}
\begingroup
\small
\setlength{\tabcolsep}{4pt}
\renewcommand{\arraystretch}{1.42}
\setlength{\heavyrulewidth}{0.65pt}
\setlength{\lightrulewidth}{0.35pt}
\arrayrulecolor{mindlabtablerule}
\captionsetup[subtable]{format=plain,labelformat=parens,labelsep=space,
  justification=raggedright,singlelinecheck=false,
  font={small,sf},labelfont={bf,color=mindlabtableink},textfont=bf,
  skip=6pt,position=top}
\begin{subtable}[t]{0.485\textwidth}
\centering
\caption{Fine-stage candidate count}
\label{tab:numerical-candidates}
\begin{tabularx}{\linewidth}{X
  >{\centering\arraybackslash}p{0.235\linewidth}
  >{\columncolor{mindlabtablehighlight}\centering\arraybackslash}p{0.235\linewidth}}
\toprule
\rowcolor{mindlabtablehead}
\parbox[c][24pt][c]{\linewidth}{\mindlabtablehead{Metric}} &
\parbox[c][24pt][c]{\linewidth}{\centering\mindlabtablehead{32 fine\\bins}} &
\parbox[c][24pt][c]{\linewidth}{\centering\mindlabtablehead{4/8 fine\\bins}} \\
\midrule
Coarse intervals & 30 & 30 \\
Total fine cells & 960 & 216 \\
Quantization MAE $\downarrow$ & \textbf{0.0391} & 0.1540 \\
Prediction MAE $\downarrow$ & 0.3277 & \textbf{0.3202} \\
\bottomrule
\end{tabularx}
\end{subtable}\hfill
\begin{subtable}[t]{0.485\textwidth}
\centering
\caption{Fine-stage decoding}
\label{tab:numerical-decoding}
\begin{tabularx}{\linewidth}{X
  >{\columncolor{mindlabtablehighlight}\centering\arraybackslash}p{0.235\linewidth}
  >{\centering\arraybackslash}p{0.275\linewidth}}
\toprule
\rowcolor{mindlabtablehead}
\parbox[c][24pt][c]{\linewidth}{\mindlabtablehead{Metric}} &
\parbox[c][24pt][c]{\linewidth}{\centering\mindlabtablehead{Argmax}} &
\parbox[c][24pt][c]{\linewidth}{\centering\mindlabtablehead{Conditional\\expectation}} \\
\midrule
MAE $\downarrow$ & \textbf{0.3202} & 0.3300 \\
AbsRel (\%) $\downarrow$ & \textbf{29.03} & 30.04 \\
$|e|\leq 0.1$ m (\%) $\uparrow$ & \textbf{59.00} & 55.67 \\
$|e|/y\leq 25\%$ (\%) $\uparrow$ & 68.12 & \textbf{70.81} \\
\bottomrule
\end{tabularx}
\end{subtable}
\endgroup
\end{table}

\paragraph{Visual refinement.}
\label{sec:visual-results}
Point prediction requires both a precise output representation and enough visual detail to distinguish nearby locations.
We investigate whether original-image context or reused features can replace fresh crop encodings while retaining localization accuracy.
Removing crops from the trained model reduces RefSpatial accuracy from 51.00\% to 9.00\%; training a model without crops recovers 46.50\%, showing the importance of adapting training to the chosen visual input (\cref{tab:nocrop,tab:roi}).
Crop re-encoding nevertheless achieves the highest RefSpatial and Point-Bench scores among the four visual strategies, reaching 51.00\% and 56.84\%, respectively.
Replaying visual tokens or attention K/V does not recover these scores.
We therefore retain crop re-encoding to supply the local visual evidence needed by successive region decisions.

We also evaluate direct visual-token pointing, which replaces the three-stage hierarchy with a single decision over the original image's visual tokens.
For the native Q/K model, connected-component decoding improves counting success from 12.76\% with argmax to 32.14\%.
Selecting the decoder according to the task raises the overall Point-Bench score from 50.76\% with argmax to 54.63\% (\cref{tab:visual-decoders}).
The hierarchical crop model retains stronger RefSpatial and Point-Bench performance (\cref{tab:visual-tokens}), while these results show how adapting the decoder makes the one-stage alternative more effective.

\begin{table}[htbp]
\centering\small
\setlength{\tabcolsep}{5pt}
\caption{Matched training for alternative visual inputs. All four models use L32 mean, 72K QA, and three nine-way decisions. Each alternative is trained from the same pretrained base.}
\label{tab:roi}
\begin{tabular}{lrrrr}
\toprule
\rowcolor{mindlabtablehead}
\mindlabtablehead{Metric} & \mindlabtablehead{Crop} & \mindlabtablehead{Original} & \mindlabtablehead{ViT replay} & \mindlabtablehead{KV replay} \\\midrule
SAT real & 78.00 & 74.67 & 73.33 & 78.00 \\
CV-Bench & 85.31 & 85.14 & 84.56 & 84.52 \\
RoboSpatial VQA & 68.06 & 68.75 & 65.01 & 63.38 \\
BLINK val & 63.04 & 62.56 & 61.07 & 62.74 \\
MMSI-Bench & 27.80 & 25.80 & 26.90 & 26.80 \\
RefSpatial & 51.00 & 46.50 & 47.00 & 45.50 \\
Where2Place & 59.98 & 60.93 & 59.04 & 68.95 \\
RoboSpatial pointing & 62.30 & 58.20 & 59.02 & 54.92 \\
Point-Bench (macro) & 56.84 & 52.55 & 52.87 & 52.76 \\
VST-300 MAE (m) & 0.3344 & 0.3577 & 0.3714 & 0.4097 \\
\bottomrule
\end{tabular}
\end{table}

\paragraph{Multi-point prediction.}
Finally, we test whether the same point-classification hierarchy can recover multiple targets without additional training.
Replacing repeated continue/stop decisions with a spatial probability tree raises Point-Bench counting success from 19.39\% to 38.78\% (\cref{tab:tree}).
Root pruning retains the same success rate while reducing the mean number of decision nodes from 91 to 51.10 and inference time from 4.716 to 2.638 seconds per question, a 1.79$\times$ speedup.
The default 10\% peak threshold achieves 38.78\% success, compared with 30.10\% and 37.24\% at thresholds of 5\% and 20\%, respectively (\cref{tab:tree-sensitivity}).
The shared classifier thus extends to multi-point prediction through its decision tree, with pruning reducing the cost of this extension.

\subsection{Training Speed}
\label{sec:cookbook}
The scheduling study tests whether similar-length microbatches and cross-GPU workload balancing reduce the cost of training heterogeneous spatial questions.
Four fresh runs use the same historical 72K corpus, global batch 256, microbatch eight, four accumulation steps, and eight L20D GPUs.
Each runs 100 updates; after eight warmup updates, the measurement window contains the same 23,552 QA.
The comparison holds the model and training computation fixed apart from scheduling.

\begin{table}[t]
\centering\small
\setlength{\tabcolsep}{5pt}
\caption{Training efficiency on the historical 72K corpus: eight GPUs, global batch 256, microbatch 8, four accumulation steps. Each condition runs 100 updates; the last 92 updates contain the same 23,552 measured QA. Tail includes communication and waiting. No quality evaluation was performed.}
\label{tab:efficiency}
\begin{tabular}{lrrrr}
\toprule
\rowcolor{mindlabtablehead}
\mindlabtablehead{Layout} & \mindlabtablehead{QA/s} & \mindlabtablehead{Padding (\%)} & \mindlabtablehead{Step CV (\%)} & \mindlabtablehead{Tail (ms)} \\\midrule
Original order & 99.06 & 35.17 & 2.17 & 39.5 \\
Local length & 116.91 & 12.36 & 4.13 & 157.6 \\
\rowcolor{mindlabtablehighlight}
Global balanced & 129.54 & 1.77 & 6.07 & 85.5 \\
Global + buffer 128 & 129.39 & 1.03 & 19.77 & 95.8 \\
\bottomrule
\end{tabular}
\end{table}

Local length grouping raises throughput from 99.06 to 116.91 QA/s and reduces padding from 35.17\% to 12.36\%, but increases mean gradient-ready spread from 115.0 to 306.6\,ms (\cref{tab:efficiency}).
Global grouping with cross-GPU balancing reduces padding to 1.77\% and ready-time spread to 145.7\,ms, reaching 129.54 QA/s: 30.78\% higher throughput, or 23.53\% less measured time for the same QA volume, than the original strategy.
Since ready-time spread remains above the original value, the gain reflects the combined scheduling change.

A 128-QA cross-update buffer lowers padding to 1.03\% but does not improve throughput and increases step-time variability.
We retain the unbuffered layout to preserve optimizer-batch membership.
Distributed checks verify per-question loss normalization with unequal workloads and incomplete tails, without implying bitwise equivalence across BF16 batch shapes.
These short runs measure throughput, not convergence or full-run duration; they include no downstream quality evaluation.
Full timing and batch-size feasibility profiles appear in \cref{app:training-measurements}.

\Needspace{16\baselineskip}
\begin{samepage}
\section{Conclusion}
MJev-Spatial investigates a System One approach to spatial understanding: using pretrained multimodal knowledge to make direct spatial decisions without generating answer text.
Options, numerical values, and image locations share a candidate-classification mechanism, with hierarchical refinement providing numerical and spatial precision and tree expansion extending point prediction without further training.
An audited 12.3M-QA corpus and joint stage supervision support training across these tasks; length grouping and cross-GPU balancing improve throughput by 30.8\% in a controlled eight-GPU study.
The final Qwen3.5-4B-based model achieves the highest overall score, leads seven of twelve reported metrics among eight evaluated models, and exceeds its local autoregressive base model on all twelve.
Its mean batch-one query latency is 160.6\,ms across the twelve benchmarks, giving 2.5$\times$ and 5.6$\times$ speedups over the fastest autoregressive baseline and the base model under the reported timing protocol.
Together with the controlled design studies, these results support direct candidate decisions as a viable approach to the spatial tasks studied here.
\par
\end{samepage}

\clearpage
\bibliographystyle{unsrtnat-sixauthors}
\bibliography{references}
\clearpage
\appendix
\section{Additional Method Details}
\label{app:implementation}
\subsection{Autoregressive Reference and Decision Paths}
\paragraph{Problem setup.}
Given ordered images $I=(I_1,\ldots,I_m)$ and a spatial question $q$, the goal is to predict a typed answer $y\in\mathcal Y_q$: one of the supplied options, a number with a declared unit, or a location in normalized image coordinates.
In an autoregressive formulation, a vision-language model $\pi_\theta$ generates an answer string $w=(w_1,\ldots,w_L)$ autoregressively,
\begin{equation}
 \pi_\theta(w\mid I,q)=\prod_{l=1}^{L}\pi_\theta(w_l\mid I,q,w_{<l}),
 \label{eq:ar}
\end{equation}
and a task-specific parser then maps $w$ to $y$.
Answer generation therefore adds sequential decoding steps whose number depends on the output length $L$.

In the classification formulation, the state at stage $t$ is $s_t=(I,q,h_{<t})$, where $h_{<t}$ contains the preceding selections and visual refinements.
A deterministic controller constructs $\mathcal C_t$, advances the state after selection, and decodes the terminal path to the required answer.
For a fixed observation and question, reachable states form a finite tree of depth at most three, with path probability
\begin{equation}
 P(a_{1:S}\mid I,q)=\prod_{t=1}^{S}p_\theta(a_t\mid I,q,h_{<t},\mathcal C_t).
 \label{eq:path}
\end{equation}
Ordinary decoding follows the greedy path in \cref{eq:choice}; multi-point decoding expands several spatial paths.
Training supervises all annotated transitions using \cref{eq:loss}.
Each textual decision supports up to 32 candidates, with padded slots masked before normalization.

\subsection{Input Sequence and Candidate Representations}
\label{app:serialization}
\paragraph{Sequence layout.}
Each question--answer example (QA) is represented by an image prefix followed by the question, candidate descriptions, and a decision marker:
\begin{equation}
 [\mathrm{state},I_1,\ldots,I_m,\mathrm{question},q,
   \mathrm{opt},c_1,\mathrm{end},\ldots,
   \mathrm{opt},c_K,\mathrm{end},\mathrm{decide}].
 \label{eq:serialization}
\end{equation}
The sequence markers reuse existing vocabulary entries, leaving the vocabulary and embedding matrix unchanged.

\paragraph{Candidate representations.}
The frozen vision encoder~\citep{vit} and merger embed each image into the language sequence, and the first 31 language layers contextualize the whole causal row.
The state at an option-end marker then summarizes that option in the context of the image, the question, and the preceding options, while the state at the decision marker sees all current candidates.
Changing an option therefore changes its representation rather than the interpretation of a fixed classifier slot, which lets the candidate set vary freely across questions.
Training shuffles candidate order, while evaluation follows each benchmark's ordering protocol.

\paragraph{Multi-stage rows.}
Each QA occupies its own causal row.
A later stage appends the selected option, an optional new image crop, and a new question and candidate block.
During training, all supervised stages fit in one teacher-forced~\citep{teacherforcing} causal sequence.
Each decision marker precedes the later choices and crops, so one forward pass supervises every stage without showing a future crop to an earlier decision.
At inference, the next stage is constructed only after the current prediction.

\subsection{Candidate Scoring with Pretrained Query and Key Projections}
\label{app:readout-details}
The shared scorer produces one class logit for each current candidate.
The final full-attention layer's pretrained query projection acts on the decision state, and its key projection acts on each candidate state; comparing the resulting vectors gives the logits used in \cref{eq:choice}.
This reuses existing parameters to compute classification scores without an additional prediction head.

\paragraph{Backbone boundary.}
Qwen3.5-4B has 32 language layers with hidden width $D=2560$.
Full attention occurs at L4, L8, \ldots, L32, and the other 24 layers are Gated DeltaNet.
We run L1--L31, including their residual and feed-forward paths, and read representations at the input to L32.
A full-attention layer has $H=16$ query heads, $G=4$ key/value heads, and head width $d=256$.

\paragraph{Query--key scoring.}
Let $u_t\in\R^{2560}$ be the L32 input state at the decision marker of stage $t$, and $v_{t,i}\in\R^{2560}$ the state at the option-end marker of candidate $i$.
With the native input RMSNorm~\citep{rmsnorm} $N_{\rm in}$, the headwise RMSNorms $N_Q,N_K$, and the partial rotary operator~\citep{rope} $R_p$, the logits are
\begin{align}
 Q_{t,h}&=R_{p_t}\!\left[N_Q\!\left(W_Q^{(h)}N_{\rm in}(u_t)\right)\right],
   &&h=0,\ldots,15,\\
 K_{t,i,g}&=R_{p_{t,i}}\!\left[N_K\!\left(W_K^{(g)}N_{\rm in}(v_{t,i})\right)\right],
   &&g=0,\ldots,3,\\
 s_{t,h,i}&=\frac{Q_{t,h}^{\top}K_{t,i,g(h)}}{\sqrt{256}},
   &&g(h)=\lfloor h/4\rfloor,\\
 z_{t,i}&=\frac1{16}\sum_{h=0}^{15}s_{t,h,i}.
 \label{eq:native}
\end{align}
The rotary positions are the original token positions of the gathered states.
Qwen's multimodal rotary embedding~\citep{qwen2vl} rotates the first 64 coordinates of each head and leaves the other 192 unchanged.
Each group of four query heads shares one key head, as in grouped-query attention~\citep{gqa}.
The score is a learned, position-aware scaled inner product.

We average the head logits and apply a single softmax over the legal candidates.
The candidate probability is \cref{eq:choice}, with only the current alternatives in the denominator.
Each logit thus decomposes into 16 head contributions, which the intervention analysis in \cref{sec:head-results} exploits.

\paragraph{Reusing the pretrained projections.}
Qwen's Q projection is a bias-free $2560\rightarrow8192$ matrix whose 512 output rows per head hold a 256-dimensional query followed by a 256-dimensional output gate.
The readout uses the 256 query dimensions of each head, giving $W_Q\in\R^{4096\times2560}$.
The key projection $W_K\in\R^{1024\times2560}$ and both head norms are kept as they are.
All of these parameters start from the official pretrained checkpoint.

\paragraph{Removing unused computation.}
Since the readout sits before the value aggregation and residual update of L32, the V projection, output-gate rows, output projection, post-attention norm, and SwiGLU FFN~\citep{swiglu} of L32, together with the final language-model norm, have no path to the loss and are removed.
Earlier layers stay intact.
The readout gathers the $S$ decision states and $\sum_tK_t$ option-end states before projecting them, so it forms neither a full-sequence Q/K projection nor an $L\times L$ attention matrix.
Its candidate-comparison work is $O(\sum_tK_tHd)$, while image encoding and the L1--L31 prefill still scale with the full input length.

The retained language model has 4,111,374,336 trainable parameters, 95,688,192 fewer than the complete-language KEV model; the removed FFN alone accounts for 70,778,880.
The Q/K projections and their head norms, 13,107,712 parameters, form the readout optimizer group, and the input norm stays in the backbone group.
Trainable weights, gradients, Adam~\citep{adam} states, and all readout operations use FP32, while the retained language layers compute in BF16, following standard mixed-precision training~\citep{mixedprecision,bfloat16}.
The vision encoder and merger stay frozen in BF16.
This keeps the small candidate comparison numerically stable while the backbone computation, which dominates the cost, runs in mixed precision.

\subsection{Multi-Point Decoding Rules}
\label{app:point-decoding}
For Point-Bench counting, the same trained model supplies a three-level conditional probability tree,
\begin{equation}
 P(a_{1:3}\mid I,q)=\prod_{t=1}^{3}
 p_\theta(a_t\mid I,q,a_{<t}),
 \label{eq:tree}
\end{equation}
whose complete expansion visits $1+9+81=91$ decision nodes and produces 729 cells.
For evaluation, we drop a first-level branch only when $p(a_1)<0.01\max_jp(j)$, so each retained root adds ten decision nodes and 81 leaf cells, and deeper levels are fully expanded.
Unvisited cells remain unknown, and the omitted mass is recorded without renormalizing the retained probabilities.
On the resulting map, cells above 10\% of the retained global peak form an eight-connected superlevel set, and each connected component outputs the center of its peak cell.
The threshold, connectivity, and root cutoff are fixed without test labels; target masks and true point counts are used only by the scorer.
This turns distributed spatial mass into a variable number of points without a learned stop decision or a ground-truth top-$k$.
The tree costs more than the three-node pointing path, and its pruning and decoder sensitivities are measured in \cref{sec:tree-results}.

\section{Data Conversion and Validation}
\label{app:data-details}
\subsection{From Spatial Annotations to Classification Supervision}
We convert spatial annotations from the training collection curated for Molmo2-ER~\citep{molmo2er}, which contains 5M source records and 21.1M QA.
The annotations originate from RefSpatial~\citep{refspatial}, SAT~\citep{sat}, SenseNova-SI~\citep{sensenova}, VSI-590K~\citep{cambrians}, VST-P~\citep{vst}, and RoboPoint~\citep{robopoint}; \cref{tab:data} reports their contributions to the final corpus.
Each accepted QA contains the ordered images, question, answer type, candidates, and target.
Supplied options retain their semantic content, numerical labels determine the target intervals, and point coordinates determine the target regions and training crops.
Source answers and rationales are excluded from model inputs.

\paragraph{Answer options.}
Supplied text and image options are retained when exactly one candidate matches the source answer.
Questions with ambiguous matches or unresolved image bindings are excluded.
Question text, including comparison clauses, is preserved separately from the candidate descriptions.

\paragraph{Numerical supervision.}
When no explicit answer menu applies, the requested quantity determines the task and units.
Length targets are converted to meters, counts are exact integers in 0--1023, and angle direction follows the question.
We accept a scalar target only when its value and unit are unambiguous.
Ranges, inequalities, conflicting quantities, and unsupported area or volume answers are excluded.

\paragraph{Point and relation supervision.}
Single-point labels with a verified coordinate convention are normalized to the original image.
Multiple-point lists, region-only annotations, unresolved image bindings, and examples with more than five source images are excluded.
Boolean and relation labels use fixed candidate vocabularies when no supplied answer menu applies.
RoboPoint-spatial~\citep{robopoint} contributes no final QA because its available multi-point or region supervision is outside this adaptation.

\subsection{Data Quality Checks}
\label{sec:data-audits}
We compare 245 converted numerical QA with their source annotations and semantically review 75 cases.
A targeted review of 66,574 QA with known option-conversion issues recovers 65,902 QA and excludes 672 ambiguous cases.

We remove adapter-added option labels and answer-format instructions while preserving the meaning of candidate text.
This cleanup affects 15.2M candidate texts across 6.5M QA; the final corpus contains 12,304,058 QA.
We check 3,422 nontrivial cleanup cases against their source annotations and validate 3,427 rendered training examples across five sources and 200 display categories, including teacher-forced crops.
A focused content review examines 393 cases.
Another 256 pointing questions are flagged for coordinate-wording review.

\subsection{Duplicate Filtering and Evaluation Separation}
We exclude invalid images and training examples whose images exactly match public evaluation images or satisfy a conservative near-duplicate criterion.
Near-duplicates require a 64-bit difference-hash~\citep{dhash} Hamming distance at most 2, aspect ratios within 1\%, and mean absolute RGB difference at most 6 on $8\times8$ thumbnails.
These checks target nearly identical images; they do not establish physical-scene or video-neighbor independence.

Image validation covers 2.1M image identities and excludes 85 images.
Conflicting targets for the same input cause all members of that group to be excluded, while identical duplicate examples are reduced to one.
The filtering stages exclude 45,820 duplicate-input QA, 3,430 QA with conflicting targets, and 201 near-duplicate QA under the public-test policy.
Examples whose estimated training sequences exceed 4,096 tokens, including refinement crops, are excluded.

\subsection{Final Corpus and One-Pass Distribution}
\Cref{tab:data} reports the source-by-task counts after conversion and option cleanup; counts denote original QA rather than decision stages.

\begin{table}[htbp]
\centering\small
\setlength{\tabcolsep}{5pt}
\caption{Final r12 corpus after content-only option cleanup. Counts denote original QA, not the number of decision stages. Source names refer to training annotations, including adaptations of existing data.}
\label{tab:data}
\begin{tabular}{lrrrrrr}
\toprule
\rowcolor{mindlabtablehead}
\mindlabtablehead{Source} & \mindlabtablehead{Choice} & \mindlabtablehead{Point} & \mindlabtablehead{Length} & \mindlabtablehead{Angle} & \mindlabtablehead{Count} & \mindlabtablehead{Total} \\\midrule
RefSpatial~\citep{refspatial} & 6.1M & 3.5M & 1.3M & 108K & 0 & 11M \\
SAT~\citep{sat} & 168K & 0 & 0 & 0 & 0 & 168K \\
SenseNova-SI~\citep{sensenova} & 424K & 0 & 34K & 2 & 25K & 484K \\
VSI-590K-image~\citep{cambrians} & 203K & 0 & 27K & 12K & 13K & 255K \\
VST-P~\citep{vst} & 218K & 0 & 170K & 0 & 0 & 388K \\
\midrule
\rowcolor{mindlabtablehighlight}
{\bfseries Total} & 7.1M & 3.5M & 1.6M & 120K & 38K & 12.3M \\
\bottomrule
\end{tabular}
\end{table}

\begin{table}[htbp]
\centering\small
\setlength{\tabcolsep}{5pt}
\caption{Natural task distribution used by the final one-pass training recipe. No task rebalancing or repeated oversampling is applied.}
\label{tab:mix}
\begin{tabular}{lrrr}
\toprule
\rowcolor{mindlabtablehead}
\mindlabtablehead{Task} & \mindlabtablehead{QA} & \mindlabtablehead{Share (\%)} & \mindlabtablehead{Decision rounds} \\\midrule
Choice & 7.1M & 57.65 & 1 \\
Point & 3.5M & 28.34 & 3 \\
Length & 1.6M & 12.71 & 1--2 \\
Angle & 120K & 0.98 & 1--2 \\
Count & 38K & 0.31 & 2 \\
\bottomrule
\end{tabular}
\end{table}

For corpus statistics, numerical supervision is separated into lengths, angles, and counts, giving five categories across the three output types.
The final 12.3M QA comprise 7.1M choices, 3.5M points, 1.6M lengths, 120K angles, and 38K counts.
RefSpatial-derived data account for approximately 89.5\% and provide all current point supervision.
The natural distribution is retained for one-pass training; no task or source is repeatedly sampled to meet a target ratio.

The length census contains 382 zeros; its median, 99th percentile, and maximum are 1.3381, 163.1174, and 8520.1\,m.
In total, 12K labels exceed the old 200\,m limit.
Count labels reach 1002 despite a 99th percentile of 49, which motivates retaining the exact 0--1023 interface.
\section{Training Configuration and Execution}
\subsection{Matched 72K Training Recipe}
The controlled studies use 72K QA: 25K choice questions from SAT~\citep{sat}, 22K length-estimation questions from VST-P~\citep{vst}, and 25K pointing questions from RefSpatial~\citep{refspatial}.
The length subset contains 11K height and 11K longest-dimension labels, ranging from 0 to 167.82\,m.
Examples, order, and one-pass exposure budget are fixed for the matched comparisons.
The aligned experiments use eight NVIDIA L20D GPUs, four QA per GPU, global batch 32, no gradient accumulation, and 2,252 optimizer updates for one pass.
For the selected native model, this pass takes 1,413\,s of training time (3.14 GPU-hours).
All three primary readout models share seed 3407, the canonical QA permutation, image budgets, stage-normalized loss, and optimizer settings.
Vision and merger parameters are frozen; retained language parameters are fully trainable.
Training allows 64--512 visual tokens per image and at most 4,096 total tokens per QA sequence.
Shared language prefixes and visual deduplication are disabled for this controlled comparison.

Fused AdamW~\citep{adamw} uses $\beta=(0.9,0.95)$, $\epsilon=10^{-8}$, global gradient-norm clipping~\citep{gradclip} at 1, and weight decay 0.01 on the backbone and 0 on the readout.
Nominal learning rates are $5\times10^{-6}$ and $3\times10^{-4}$.
With zero-based update $u$ and horizon $T$, the implemented multipliers combine linear warmup~\citep{warmup} with cosine decay~\citep{sgdr}:
\begin{align}
 d(u)&=0.1+0.9\frac{1+\cos(\pi u/T)}{2},\\
 \eta_{\mathrm{head}}(u)&=\eta_{\mathrm{head}}^{\max}
       \min(1,(u+1)/100)\,d(u),\\
 \eta_{\mathrm{backbone}}(u)&=\eta_{\mathrm{backbone}}^{\max}
       \min(1,\max(0,u+1-20)/100)\,d(u).
\end{align}
The backbone learning rate is zero for the first 20 updates, while Adam moments continue to update.

\subsection{Large-Scale Training Configuration}
\label{sec:formal}
\begin{table}[htbp]
\centering\small
\caption{Recorded configuration for training on the 12.3M-QA corpus.}
\label{tab:formal}
\begin{tabular}{ll}
\toprule
\rowcolor{mindlabtablehead}
\mindlabtablehead{Item} & \mindlabtablehead{Configuration} \\\midrule
Initialization & Official post-trained Qwen3.5-4B; fresh from base\\
Readout & L32 native Q/K mean; no layer fusion\\
Data & r12 content-only; 12.3M QA; each exactly once\\
Global / per-GPU micro / accumulation & 512 / 8 / 8 on eight GPUs\\
Updates / last real batch & 24,032 / 186 QA\\
Backbone / readout LR & $5\times10^{-6}$ / $2\times10^{-4}$\\
Optimizer and schedule & AdamW and the same update-based 20/100 schedule\\
Precision & BF16 compute; FP32 language weights, gradients, Adam and readout\\
Length / pointing & 30 coarse intervals, 4/8 fine choices / three pointing stages\\
Input limits & 5 source images; 64--512 visual tokens/image; 4,096 tokens\\
Batch layout & Global length grouping and rank balancing; no buffer\\
Task sampling & One flat permutation, seed 3407; no task resampling\\
\bottomrule
\end{tabular}
\end{table}

A complete one-pass run under this configuration consists of 24,032 updates and 12,304,058 real QA exposures.

\subsection{Scheduling and Distributed Execution}
\label{app:training-execution}
To reduce padding, sequences within each global optimizer batch are sorted by expanded length $L_q$ into microbatches of eight.
To limit workload imbalance across devices, a longest-work-first assignment~\citep{lpt} distributes these microbatches across eight ranks by the cost
\begin{equation}
 C(B)=|B|\max_{q\in B}L_q+0.25\sum_{q\in B}V_q,
 \label{eq:batch-cost}
\end{equation}
where $V_q$ measures visual-token work, and each scheduling round gives one microbatch to each rank.
The recipe uses no cross-update buffer, concatenated QA sequences, training prefix cache, or image deduplication.

With $W$ workers, each microbatch uses $W/N_{\rm actual}$ times its weighted loss sum, so averaging across workers~\citep{pytorchddp} recovers \cref{eq:loss}.

\section{Evaluation Protocols}
\label{app:evaluation-protocols}
\subsection{72K Pilot Evaluation Scope and Scoring}
The historical 72K pilot evaluation union contains eight public datasets, represented as 7.5K execution requests (\cref{tab:benchmarks}).
SAT contributes 150 real-image questions evaluated in both answer orders, so the union has 150 fewer independent questions than requests.
The historical program additionally evaluates VST-300, a fixed 300-example internal development set for length prediction.

\begin{table}[htbp]
\centering\small
\caption{Historical 72K pilot evaluation scope.
Counts are execution requests.}
\label{tab:benchmarks}
\begin{tabularx}{\textwidth}{@{}l r X@{}}
\toprule
\rowcolor{mindlabtablehead}
\mindlabtablehead{Benchmark} & \mindlabtablehead{Requests} & \mindlabtablehead{Score and protocol} \\\midrule
SAT real~\citep{sat} & 300 & Mean accuracy over both orders of 150 real-image questions.\\
CV-Bench~\citep{cvbench} & 2.6K & $0.25\,\mathrm{ADE20K}+0.25\,\mathrm{COCO}+0.5\,\mathrm{Omni3D}$~\citep{ade20k,coco,omni3d};
counting is included in the overall score.\\
RoboSpatial VQA~\citep{robospatial} & 228 & Mean of configuration (123) and compatibility (105) accuracies;
RGB input with the current annotations.\\
BLINK val~\citep{blink} & 1.9K & Macro average of all 14 public validation tasks; no test split.
Use the paired Hugging Face options and labels.\\
MMSI-Bench~\citep{mmsi} & 1K & Exact-option accuracy over all 11 types; all 2--10 images in
original order; not circular accuracy.\\
RefSpatial~\citep{refspatial} & 200 & Binary-mask mean point accuracy; location 100 and placement 100;
the overlapping unseen subset is not added again.\\
Where2Place~\citep{robopoint} & 100 & Grayscale JPEG mask value divided by 255, averaged over predicted
points and then questions.\\
RoboSpatial pointing~\citep{robospatial} & 122 & Current author scorer: any of the first two predicted
points hits the mask; our grid decoder emits one.\\
Point-Bench~\citep{pointarena} & 982 & Macro average of five category success rates; counting also
requires exact point count. Four non-counting categories: 786; counting: 196.\\
\bottomrule
\end{tabularx}
\end{table}

Inference receives images, questions, and declared answer spaces only.
Ground-truth values, evaluation masks, and target object counts are used by offline scoring.
They do not select crops, tree branches, component thresholds, or output point counts.
Original benchmark options and image order are retained, except for SAT's explicit two-order protocol.
Invalid outputs remain in the denominator.
Point-Bench's three recorded image/mask dimension anomalies are retained rather than dropped.

There are two important historical measurement distinctions.
First, the initial matched-head evaluation used an 8,192-token sequence cap; the extended benchmarks and later matched reruns use 32,768, with up to 1,024 visual tokens per image throughout quality evaluation.
The consolidated historical matrix retains each recorded run and its original evaluation budget.
Second, earlier Point-Bench summaries used the 982-question micro average.
This report consistently computes the five-category macro average from retained category scores.

\subsection{Local Autoregressive Baseline for the 72K Pilot}
The unmodified post-trained Qwen3.5-4B checkpoint is evaluated locally with BF16, native Transformers~\citep{transformers} generation, PyTorch SDPA~\citep{pytorch}, greedy decoding, one beam, and thinking disabled.
Its limits are 2,048 generated tokens for choice and pointing, 128 for numerical answers, and 32,768 total tokens with at most 1,024 visual tokens per image.
Choice answers are parsed as option identifiers, and point predictions use integer coordinates in $[0,1000]$.
The native model can emit multiple points.
Inputs are not silently truncated and incorrect or invalid cases are not retried according to their labels.

\subsection{Final Inference-Efficiency Protocol}
\label{app:final-latency}
\Cref{tab:final-inference-efficiency} reports mean latency over the complete evaluation sets of all twelve benchmarks.
Each benchmark contributes equally to Overall, irrespective of its question count.
The eight models share one NVIDIA B200, batch size one, and the reported end-to-end timing boundary.
\begin{table}[htbp]
\centering
\footnotesize
\setlength{\tabcolsep}{3.0pt}
\renewcommand{\tabularxcolumn}[1]{m{#1}}
\begin{threeparttable}
\caption{Final-model inference efficiency on the complete twelve-benchmark evaluation sets. Entries are mean single-query latency in milliseconds on one NVIDIA B200, batch size one. Timing includes image I/O, preprocessing, host-to-device transfer, inference/decision, and decoding; warmup and model loading are excluded. The two panels contain the same eight models.}
\label{tab:final-inference-efficiency}
\begin{tabularx}{\textwidth}{@{}>{\raggedright\arraybackslash}m{.26\textwidth}*{6}{>{\raggedleft\arraybackslash}X}@{}}
\toprule
\rowcolor{mindlabtablehead}
\mindlabtablehead{Model} & \mindlabtablehead{SAT} & \mindlabtablehead{CV-Bench} & \mindlabtablehead{Robo-VQ} & \mindlabtablehead{BLINK} & \mindlabtablehead{MMSI} & \mindlabtablehead{RefSpatial} \\\midrule
\rowcolor{mindlabtablehighlight}
{\bfseries MJev-Spatial} & 155.1 & 76.5 & 88.9 & 106.6 & 116.0 & 204.8 \\
\shortstack[l]{Spatial-SSRL-\\Qwen3VL-4B} & 237.8 & 127.7 & 123.8 & 124.0 & 139.4 & 682.0 \\
SpatialThinker-7B & 266.4 & 120.8 & 152.3 & 172.4 & 236.8 & 458.8 \\
\shortstack[l]{Qwen3-VL-8B-\\Instruct} & 250.8 & 128.5 & 111.9 & 128.6 & 151.0 & 855.5 \\
\shortstack[l]{SenseNova-SI\\Qwen3-VL-8B} & 223.3 & 115.4 & 102.7 & 105.0 & 114.0 & 858.7 \\
Molmo2-ER & 343.3 & 146.5 & 226.1 & 280.3 & 348.5 & 373.3 \\
Qwen3.5-4B & 364.4 & 241.9 & 201.9 & 212.4 & 226.9 & 1490.9 \\
InternVL3.5-8B & 388.0 & 250.7 & 214.0 & 234.8 & 325.4 & 2928.4 \\
\bottomrule
\end{tabularx}

\vspace{0.8em}
\begin{tabularx}{\textwidth}{@{}>{\raggedright\arraybackslash}m{.26\textwidth}*{8}{>{\raggedleft\arraybackslash}X}@{}}
\toprule
\rowcolor{mindlabtablehead}
\mindlabtablehead{Model} & \mindlabtablehead{\shortstack{Where\\2Place}} & \mindlabtablehead{\shortstack{Robo-\\POI}} & \mindlabtablehead{\shortstack{Point-\\Bench}} & \mindlabtablehead{ERQA} & \mindlabtablehead{\shortstack{Emb\\Spatial}} & \mindlabtablehead{\shortstack{VSI\\full}} & \mindlabtablehead{\shortstack{Overall\\(ms)}} & \mindlabtablehead{\shortstack{Load\\(s)}} \\\midrule
\rowcolor{mindlabtablehighlight}
{\bfseries MJev-Spatial} & 195.4 & 264.6 & 301.2 & 80.0 & 69.9 & 268.2 & {\bfseries 160.6} & 14.4 \\
\shortstack[l]{Spatial-SSRL-\\Qwen3VL-4B} & 681.0 & 701.4 & 1193.4 & 84.4 & 77.5 & 596.5 & \underline{397.4} & 10.5 \\
SpatialThinker-7B & 540.6 & 604.2 & 774.5 & 111.9 & 99.1 & 1343.1 & 406.7 & 11.3 \\
\shortstack[l]{Qwen3-VL-8B-\\Instruct} & 647.5 & 719.2 & 1469.6 & 91.1 & 82.2 & 668.1 & 442.0 & 12.7 \\
\shortstack[l]{SenseNova-SI\\Qwen3-VL-8B} & 646.6 & 733.1 & 5513.4 & 106.1 & 79.7 & 401.1 & 749.9 & 10.8 \\
Molmo2-ER & 2245.8 & 514.3 & 772.1 & 164.5 & 130.5 & 3782.2 & 777.3 & 10.6 \\
Qwen3.5-4B & 1451.1 & 1427.1 & 2661.5 & 180.8 & 175.0 & 2095.2 & 894.1 & 11.8 \\
InternVL3.5-8B & 7958.5 & 9411.5 & 2369.4 & 177.4 & 382.0 & 15780.7 & 3368.4 & 10.8 \\
\bottomrule
\end{tabularx}
\begin{tablenotes}[flushleft]
\footnotesize
\item Overall is the equal-weight mean of the twelve benchmark mean latencies. MJev-Spatial takes 160.6\,ms, compared with 397.4\,ms for the fastest baseline, Spatial-SSRL-Qwen3VL-4B, and 894.1\,ms for Qwen3.5-4B: latency ratios of 2.47$\times$ and 5.57$\times$, respectively.
\item MJev-Spatial's VSI profile uses 32 frames and 64 image tokens per frame. Point-Bench combines single-point and counting queries. Model-load time is reported separately.
\end{tablenotes}
\end{threeparttable}
\end{table}

\section{Extended Readout Studies}
\label{app:readout-studies}
\subsection{Readout Controls and Optimization Alignment}
\label{app:readout-controls}
We test whether a pretrained candidate scorer can replace an added classification head while retaining competitive spatial performance.
The controlled comparison evaluates three ways to convert contextual representations into candidate scores.
A \emph{position classifier} applies LayerNorm~\citep{layernorm} and a 32-output linear layer to the decision state; its outputs refer to candidate slots.
A \emph{KEV pointer} completes the language backbone and projects the decision and option-end states with two newly initialized affine $2560\rightarrow256$ maps, scoring their dot product divided by $\sqrt{256}$.
It adds 1,311,232 parameters and explicitly represents candidate content.
The \emph{native mean} model instead reuses the final attention projections at the earlier boundary in \cref{sec:readout}.
The comparison therefore tests both representation source and the amount of downstream computation retained.

\Cref{tab:aligned} is the primary matched readout comparison.
Native mean improves over the linear classifier by 1.58 points on CV-Bench, 7.72 on RoboSpatial VQA, and 2.00 on RefSpatial; MAE falls by 8.93\%.
The gains come with losses of 8.01 points on Where2Place and 3.28 on RoboSpatial pointing.
Relative to KEV, native mean improves RefSpatial and length error but is not uniformly better on choice tasks.
Its retained parameter count and measured training cost are lower.
These results support pretrained candidate scoring as a compact shared design; they do not establish that its scoring formula alone causes every gain, since templates and initialization source remain architecture-dependent.

\subsection{Pilot Quality and Task-Specific Failures}
\label{app:pilot-quality}
\begin{table}[htbp]
\centering\small
\setlength{\tabcolsep}{5pt}
\caption{Completed Qwen experiments on the fixed evaluation union. All adapted models use the 72K-QA recipe and the historical length codebook; these are not the 12.3M-QA model. Higher is better except VST MAE. Point-Bench is recomputed as the five-category macro average from retained category scores. Original run provenance and differing sequence budgets are retained.}
\label{tab:main}
\begin{tabular}{lrrrrrr}
\toprule
\rowcolor{mindlabtablehead}
\mindlabtablehead{Metric} & \mindlabtablehead{AR} & \mindlabtablehead{Linear} & \mindlabtablehead{KEV} & \mindlabtablehead{Mean} & \mindlabtablehead{Concat} & \mindlabtablehead{Merge} \\\midrule
SAT real & 43.00 & 78.33 & 76.33 & 78.00 & 75.67 & 72.00 \\
CV-Bench & 85.56 & 83.80 & 85.31 & 85.38 & 85.12 & 85.22 \\
RoboSpatial VQA & 56.64 & 60.34 & 68.54 & 68.06 & 66.84 & 65.27 \\
BLINK val & 62.98 & 61.71 & 62.99 & 63.04 & 62.69 & 61.97 \\
MMSI-Bench & 34.20 & 27.70 & 25.10 & 27.70 & 28.90 & 27.10 \\
RefSpatial & 46.00 & 49.00 & 46.00 & 51.00 & 51.00 & 50.00 \\
Where2Place & 55.00 & 67.99 & 61.02 & 59.98 & 60.05 & 63.06 \\
RoboSpatial pointing & 50.00 & 65.57 & 63.11 & 62.30 & 61.48 & 63.93 \\
Point-Bench (macro) & 62.75 & 57.55 & 57.33 & 56.74 & 55.39 & 56.40 \\
VST-300 MAE (m) & 0.6883 & 0.3668 & 0.3614 & 0.3341 & 0.3297 & 0.3405 \\
\bottomrule
\end{tabular}
\end{table}

The historical 72K evaluation union in \cref{tab:main} compares the selected readout with the unmodified local AR baseline and the native aggregation variants.
Native mean is close to AR on CV-Bench (85.38 versus 85.56) and BLINK (63.04 versus 62.98), and improves RefSpatial (51.00 versus 46.00) and RoboSpatial pointing (62.30 versus 50.00).

Category-level results expose structure that the aggregate scores conceal (\cref{tab:categories}).
On RefSpatial, the native model improves placement from 29.00 to 47.00 but reduces location from 63.00 to 55.00.
On RoboSpatial VQA, compatibility rises from 35.24 to 52.38, compared with a smaller configuration gain from 78.05 to 83.74.
Nearly equal BLINK macro scores also hide large offsets: multi-view reasoning improves by 33.08 points, while counting decreases by 29.17.
MMSI remains weak overall (27.70 versus AR's 34.20), with particularly large losses on measurement attributes and camera--object relationships.

Point-Bench is especially diagnostic.
Native mean's counting success is 38.78 versus AR's 66.84.
This difference accounts for 5.61 of the 6.01-point macro gap; the mean across the other four categories differs by only about 0.50 points.
The shortfall is concentrated in estimating and separating multiple targets, rather than a uniform loss of single-point localization.
The training mixture provides mostly one point per example, and the multi-point decoder is composed from conditional single-point decisions.

\paragraph{Complete category results.}
\label{app:categories}
The category table preserves the task structure of the six historical pilot models; it is not a category breakdown of the final release.
CV counting remains included in the official overall score and in the machine-readable export, without a separate display row here.
\begingroup
\mindlabtablestyle
\small
\setlength{\tabcolsep}{4pt}
\begin{longtable}{@{}p{.34\textwidth}rrrrrr@{}}
\caption{Category-level main comparison. Percentages use each task\textquotesingle s recorded scorer. Historical length codebook and root-pruned Point-Bench. Categories do not share a common difficulty scale.}\label{tab:categories}\\
\toprule
\rowcolor{mindlabtablehead}
\mindlabtablehead{Category} & \mindlabtablehead{AR} & \mindlabtablehead{Linear} & \mindlabtablehead{KEV} & \mindlabtablehead{Mean} & \mindlabtablehead{Concat} & \mindlabtablehead{Merge} \\\midrule\endfirsthead
\caption[]{Category-level main comparison (continued).}\\
\toprule
\rowcolor{mindlabtablehead}
\mindlabtablehead{Category} & \mindlabtablehead{AR} & \mindlabtablehead{Linear} & \mindlabtablehead{KEV} & \mindlabtablehead{Mean} & \mindlabtablehead{Concat} & \mindlabtablehead{Merge} \\\midrule\endhead
\midrule
\multicolumn{7}{r}{\footnotesize\color{mindlabmuted}Continued on next page}\\
\endfoot
\bottomrule\endlastfoot
\rowcolor{mindlabtablehighlight}
\multicolumn{7}{l}{\mindlabtablesection{SAT real}}\\*
action\_conseq & 74.32 & 83.78 & 78.38 & 78.38 & 75.68 & 77.03 \\*
ego\_movement & 8.70 & 95.65 & 100.00 & 100.00 & 100.00 & 67.39 \\*
goal\_aim & 48.53 & 95.59 & 92.65 & 94.12 & 91.18 & 94.12 \\*
obj\_movement & 41.30 & 76.09 & 73.91 & 73.91 & 65.22 & 69.57 \\*
perspective & 27.27 & 43.94 & 42.42 & 48.48 & 50.00 & 48.48 \\
\rowcolor{mindlabtablehighlight}
\multicolumn{7}{l}{\mindlabtablesection{CV-Bench}}\\*
Depth & 94.17 & 93.50 & 95.83 & 96.00 & 95.50 & 96.50 \\*
Distance & 89.00 & 91.00 & 91.50 & 89.67 & 91.00 & 89.83 \\*
Relation & 94.00 & 94.92 & 93.85 & 94.46 & 94.46 & 93.54 \\
\rowcolor{mindlabtablehighlight}
\multicolumn{7}{l}{\mindlabtablesection{RoboSpatial VQA}}\\*
compatibility & 35.24 & 52.38 & 53.33 & 52.38 & 52.38 & 47.62 \\*
configuration & 78.05 & 68.29 & 83.74 & 83.74 & 81.30 & 82.93 \\
\rowcolor{mindlabtablehighlight}
\multicolumn{7}{l}{\mindlabtablesection{BLINK val}}\\*
Art\_Style & 67.52 & 69.23 & 71.79 & 73.50 & 79.49 & 74.36 \\*
Counting & 71.67 & 54.17 & 46.67 & 42.50 & 45.83 & 43.33 \\
Forensic\_Detection & 79.55 & 61.36 & 76.52 & 65.91 & 62.12 & 61.36 \\
Functional\_Correspondence & 43.08 & 36.92 & 39.23 & 30.77 & 26.15 & 39.23 \\
IQ\_Test & 22.67 & 30.00 & 24.67 & 32.67 & 31.33 & 28.00 \\
Jigsaw & 61.33 & 55.33 & 60.67 & 67.33 & 58.67 & 57.33 \\
Multi-view\_Reasoning & 45.86 & 88.72 & 74.44 & 78.95 & 90.23 & 87.97 \\
Object\_Localization & 66.39 & 49.18 & 55.74 & 51.64 & 55.74 & 51.64 \\
Relative\_Depth & 85.48 & 87.10 & 88.71 & 82.26 & 86.29 & 84.68 \\
Relative\_Reflectance & 29.85 & 47.01 & 55.97 & 56.72 & 55.22 & 45.52 \\
Semantic\_Correspondence & 50.36 & 43.88 & 45.32 & 48.20 & 38.13 & 49.64 \\
Spatial\_Relation & 83.22 & 76.92 & 76.92 & 79.72 & 79.72 & 78.32 \\
Visual\_Correspondence & 89.53 & 93.02 & 88.95 & 92.44 & 89.53 & 88.37 \\
Visual\_Similarity & 85.19 & 71.11 & 76.30 & 80.00 & 79.26 & 77.78 \\
\rowcolor{mindlabtablehighlight}
\multicolumn{7}{l}{\mindlabtablesection{MMSI-Bench}}\\*
Attribute (Appr.) & 27.27 & 16.67 & 10.61 & 18.18 & 18.18 & 16.67 \\*
Attribute (Meas.) & 37.50 & 25.00 & 28.12 & 23.44 & 26.56 & 29.69 \\
MSR & 28.28 & 25.25 & 23.23 & 25.25 & 23.74 & 26.26 \\
Motion (Cam.) & 31.08 & 32.43 & 29.73 & 32.43 & 37.84 & 44.59 \\
Motion (Obj.) & 27.63 & 25.00 & 23.68 & 28.95 & 26.32 & 26.32 \\
Positional Relationship (Cam.--Cam.) & 34.41 & 29.03 & 29.03 & 23.66 & 24.73 & 21.51 \\
Positional Relationship (Cam.--Obj.) & 41.86 & 33.72 & 31.40 & 25.58 & 41.86 & 29.07 \\
Positional Relationship (Cam.--Reg.) & 44.58 & 32.53 & 26.51 & 39.76 & 38.55 & 22.89 \\
Positional Relationship (Obj.--Obj.) & 28.72 & 29.79 & 23.40 & 26.60 & 25.53 & 27.66 \\
Positional Relationship (Obj.--Reg.) & 43.53 & 28.24 & 21.18 & 32.94 & 30.59 & 27.06 \\
Positional Relationship (Reg.--Reg.) & 38.27 & 27.16 & 29.63 & 29.63 & 29.63 & 28.40 \\
\rowcolor{mindlabtablehighlight}
\multicolumn{7}{l}{\mindlabtablesection{RefSpatial}}\\*
location & 63.00 & 49.00 & 47.00 & 55.00 & 55.00 & 57.00 \\*
placement & 29.00 & 49.00 & 45.00 & 47.00 & 47.00 & 43.00 \\
\rowcolor{mindlabtablehighlight}
\multicolumn{7}{l}{\mindlabtablesection{Where2Place}}\\*
seen & 52.86 & 68.59 & 62.94 & 60.02 & 62.93 & 62.96 \\*
unseen & 60.00 & 66.60 & 56.55 & 59.90 & 53.33 & 63.31 \\
\rowcolor{mindlabtablehighlight}
\multicolumn{7}{l}{\mindlabtablesection{Point-Bench (macro)}}\\*
affordable & 74.24 & 77.78 & 78.79 & 78.79 & 81.82 & 80.81 \\*
counting & 66.84 & 37.76 & 38.78 & 38.78 & 34.69 & 34.18 \\*
reasoning & 67.36 & 59.07 & 56.48 & 60.10 & 53.89 & 55.44 \\*
spatial & 71.79 & 65.13 & 65.13 & 61.03 & 63.08 & 63.59 \\*
steerable & 33.50 & 48.00 & 47.50 & 45.00 & 43.50 & 48.00 \\
\end{longtable}
\endgroup

\FloatBarrier

\subsection{Readout Depth and Head Aggregation}
\label{sec:depth-results}
\begin{samepage}
\paragraph{Alternatives.}
We train mean, concat, and merged-Q/K readouts at each full-attention depth L4, L8, L12, L16, L20, L24, L28, and L32.
The selected layer's downstream branches and all later layers are removed.
L32 mean reuses the matched checkpoint; the other 23 models start independently from the same base.
The same 72K QA, 2,252 updates, optimizer groups, and evaluation protocol apply.
Retained language parameters range from approximately 0.988 billion at L4 to 4.111 billion at L32, so this is an equal-data rather than equal-compute comparison.
\par
\end{samepage}

Concat flattens the normalized, rotated head features and scales by $\sqrt{Hd}$.
For fixed weights,
\begin{equation}
 z_i^{\rm concat}=\frac{\sum_h Q_h^\top K_{i,g(h)}}{\sqrt{Hd}}
 =\sqrt H\,z_i^{\rm mean}=4z_i^{\rm mean}.
 \label{eq:concat}
\end{equation}
It changes softmax sharpness and CE gradients; it introduces no cross-head products.
Merge instead averages pretrained projection weights in FP32 \emph{before} the native head normalizations, retaining one 256-dimensional Q/K pair.
This reduces the readout to 1,311,232 parameters, but is not the same function as averaging normalized head scores.

\begin{figure}[H]
\centering
\includegraphics[width=\textwidth]{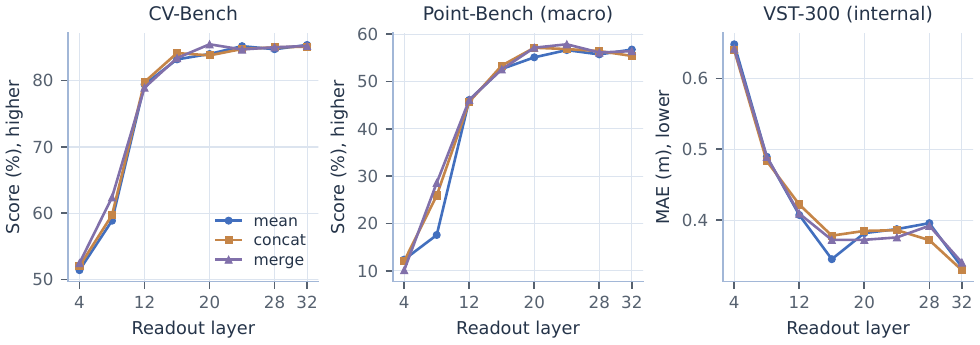}
\caption{\captionlead{The complete $8\times3$ depth/aggregation experiment, shown on three representative metrics.}
Early visual readouts are inadequate, while later layers exhibit task-dependent rather than monotonic gains.
All 24 models' benchmark values are provided in \cref{app:sweep}.}
\label{fig:sweep}
\end{figure}

\paragraph{What depth contributes.}
At L4, mean obtains 51.41 CV-Bench and 4.50 RefSpatial; at L12 these rise to 79.42 and 37.00, and at L32 to 85.38 and 51.00.
Substantial contextual processing is therefore necessary before the pretrained query and key projections provide an effective spatial readout.
Yet the optimum is task-dependent.
L28 mean reaches 53.00 RefSpatial, L24 concat reaches 68.07 Where2Place, and L28 merge reaches 67.21 RoboSpatial pointing.
Length error is also nonmonotonic: L16 mean already achieves 0.3452\,m, worsening through L20--L28 before recovering to 0.3341\,m at L32.
A high shallow-layer score on one task, such as RoboSpatial VQA, coexists with weak geometric localization.

\paragraph{Why keep L32 mean.}
At L32, concat has slightly lower MAE than mean (0.3297 versus 0.3341\,m) and better MMSI (28.90 versus 27.70), but lower SAT, RoboSpatial VQA, BLINK, and Point-Bench.
Merge improves two placement-oriented point scores relative to mean, while reducing several choice scores and worsening MAE.
Mean preserves all pretrained head subspaces without a sharper fixed logit scale or weight-space averaging.
We retain L32 mean as a balanced default with full contextual depth.

\paragraph{Late-layer fusion.}
\begin{table}[htbp]
\centering\small
\setlength{\tabcolsep}{5pt}
\caption{Equal-weight native Q/K layer fusion. Each fusion is trained independently; L32 is re-evaluated with the same evaluator. Point-Bench uses the five-category macro average.}
\label{tab:fusion}
\begin{tabular}{lrrrr}
\toprule
\rowcolor{mindlabtablehead}
\mindlabtablehead{Metric} & \mindlabtablehead{L32} & \mindlabtablehead{+L28} & \mindlabtablehead{+L24} & \mindlabtablehead{+L28+L24} \\\midrule
SAT real & 78.00 & 75.33 & 76.33 & 78.67 \\
CV-Bench & 85.31 & 84.66 & 83.46 & 85.31 \\
RoboSpatial VQA & 68.06 & 67.18 & 66.70 & 65.62 \\
BLINK val & 63.04 & 62.25 & 63.37 & 62.03 \\
MMSI-Bench & 27.80 & 27.00 & 27.00 & 27.70 \\
RefSpatial & 51.00 & 47.50 & 50.00 & 48.00 \\
Where2Place & 59.98 & 65.98 & 63.02 & 62.97 \\
RoboSpatial pointing & 62.30 & 60.66 & 60.66 & 63.11 \\
Point-Bench (macro) & 56.84 & 56.53 & 55.90 & 57.25 \\
VST-300 MAE (m) & 0.3344 & 0.3875 & 0.3993 & 0.3550 \\
\bottomrule
\end{tabular}
\end{table}

Equal-weight late-layer fusion tests whether task-specific strengths can be combined without adding projections.
For $\mathcal S\in\{\{32,28\}, \{32,24\},\{32,28,24\}\}$,
\begin{equation}
 z_i=|\mathcal S|^{-1}\sum_{\ell\in\mathcal S}z_i^{(\ell)}.
\end{equation}
An intermediate layer is executed normally in the backbone, while its existing input norm and Q/K weights also score candidates at that boundary.
Only the final fused logits receive CE.
Intermediate weights keep their backbone learning rate; the L32 readout retains its readout rate.
There are no trainable fusion gates, new parameter copies, or auxiliary losses.
The L32-only checkpoint is reevaluated with the same program as the three newly trained fusions.

\Cref{tab:fusion} shows complementary information but incomplete transfer.
L32+L28 raises Where2Place from 59.98 to 65.98, yet length MAE worsens from 0.3344 to 0.3875\,m.
All three fusions worsen MAE; the three-layer variant also reduces RoboSpatial VQA from 68.06 to 65.62.
Thus late-layer averaging does not consistently combine the individual layers' strengths under a single loss.
The final model retains one readout layer, leaving task-conditioned or learned fusion as untested alternatives.

\paragraph{Complete sweep results.}
\label{app:sweep}
Each depth retains the original layers before the selected full-attention layer and that layer's native Q/K readout.
All retained language parameters are trained.
L32 mean is reused from the aligned study, and 23 configurations are newly trained.
\begingroup
\mindlabtablestyle
\small
\setlength{\tabcolsep}{5pt}
\setlength{\LTpre}{8pt}
\setlength{\LTpost}{8pt}
\begin{longtable}{@{}lrrrrr@{}}
\caption{Complete depth/readout sweep: classification metrics. Every model uses the same QA budget and aligned recipe. The retained parameter count varies with depth. Point-Bench is macro.}\label{tab:sweep-classification}\\
\toprule
\rowcolor{mindlabtablehead}
\mindlabtablehead{Readout} & \mindlabtablehead{SAT} & \mindlabtablehead{CV} & \mindlabtablehead{Robo VQA} & \mindlabtablehead{BLINK} & \mindlabtablehead{MMSI} \\\midrule
\endfirsthead
\caption[]{Complete depth/readout sweep: classification metrics (continued).}\\
\toprule
\rowcolor{mindlabtablehead}
\mindlabtablehead{Readout} & \mindlabtablehead{SAT} & \mindlabtablehead{CV} & \mindlabtablehead{Robo VQA} & \mindlabtablehead{BLINK} & \mindlabtablehead{MMSI} \\\midrule
\endhead
\midrule
\multicolumn{6}{r}{\footnotesize\color{mindlabmuted}Continued on next page}\\
\endfoot
\bottomrule\endlastfoot
L4 mean & 52.67 & 51.41 & 72.32 & 40.34 & 27.00 \\*
L4 concat & 49.67 & 52.02 & 64.51 & 41.46 & 26.30 \\*
L4 merge & 49.67 & 52.48 & 71.03 & 41.24 & 28.30 \\
\addlinespace[2pt]
L8 mean & 49.67 & 58.87 & 66.07 & 41.37 & 25.10 \\*
L8 concat & 53.33 & 59.71 & 50.38 & 39.58 & 26.20 \\*
L8 merge & 52.33 & 62.39 & 73.61 & 40.65 & 24.50 \\
\addlinespace[2pt]
L12 mean & 61.33 & 79.42 & 51.87 & 49.49 & 28.00 \\*
L12 concat & 54.00 & 79.79 & 66.20 & 51.95 & 25.70 \\*
L12 merge & 56.33 & 78.93 & 67.49 & 52.56 & 25.80 \\
\addlinespace[2pt]
L16 mean & 74.33 & 83.20 & 63.79 & 54.96 & 27.40 \\*
L16 concat & 76.00 & 84.15 & 59.78 & 56.52 & 26.00 \\*
L16 merge & 70.67 & 83.40 & 65.82 & 57.98 & 27.00 \\
\addlinespace[2pt]
L20 mean & 77.67 & 84.01 & 64.00 & 61.97 & 27.00 \\*
L20 concat & 76.00 & 83.81 & 64.94 & 62.60 & 26.50 \\*
L20 merge & 74.00 & 85.48 & 68.75 & 61.24 & 28.60 \\
\addlinespace[2pt]
L24 mean & 76.33 & 85.20 & 61.41 & 59.98 & 25.90 \\*
L24 concat & 68.67 & 84.77 & 65.62 & 59.12 & 29.30 \\*
L24 merge & 79.00 & 84.69 & 63.59 & 64.29 & 27.30 \\
\addlinespace[2pt]
L28 mean & 72.33 & 84.74 & 65.96 & 59.98 & 26.20 \\*
L28 concat & 75.00 & 85.02 & 66.56 & 62.63 & 28.00 \\*
L28 merge & 74.67 & 85.03 & 67.79 & 61.58 & 26.50 \\
\addlinespace[2pt]
L32 mean & 78.00 & 85.38 & 68.06 & 63.04 & 27.70 \\*
L32 concat & 75.67 & 85.12 & 66.84 & 62.69 & 28.90 \\*
L32 merge & 72.00 & 85.22 & 65.27 & 61.97 & 27.10 \\
\end{longtable}
\endgroup

\begingroup
\mindlabtablestyle
\small
\setlength{\tabcolsep}{5pt}
\setlength{\LTpre}{8pt}
\setlength{\LTpost}{8pt}
\begin{longtable}{@{}lrrrrr@{}}
\caption{Complete depth/readout sweep: pointing metrics. Every model uses the same QA budget and aligned recipe. The retained parameter count varies with depth. Point-Bench is macro.}\label{tab:sweep-pointing}\\
\toprule
\rowcolor{mindlabtablehead}
\mindlabtablehead{Readout} & \mindlabtablehead{RefSpatial} & \mindlabtablehead{W2P} & \mindlabtablehead{Robo point} & \mindlabtablehead{PB macro} & \mindlabtablehead{MAE (m)} \\\midrule
\endfirsthead
\caption[]{Complete depth/readout sweep: pointing metrics (continued).}\\
\toprule
\rowcolor{mindlabtablehead}
\mindlabtablehead{Readout} & \mindlabtablehead{RefSpatial} & \mindlabtablehead{W2P} & \mindlabtablehead{Robo point} & \mindlabtablehead{PB macro} & \mindlabtablehead{MAE (m)} \\\midrule
\endhead
\midrule
\multicolumn{6}{r}{\footnotesize\color{mindlabmuted}Continued on next page}\\
\endfoot
\bottomrule\endlastfoot
L4 mean & 4.50 & 13.96 & 4.10 & 12.38 & 0.6480 \\*
L4 concat & 3.00 & 13.96 & 4.92 & 12.07 & 0.6397 \\*
L4 merge & 6.50 & 12.96 & 5.74 & 10.15 & 0.6417 \\
\addlinespace[2pt]
L8 mean & 7.00 & 14.95 & 17.21 & 17.58 & 0.4895 \\*
L8 concat & 15.00 & 29.93 & 28.69 & 25.92 & 0.4833 \\*
L8 merge & 16.00 & 33.91 & 31.15 & 28.58 & 0.4894 \\
\addlinespace[2pt]
L12 mean & 37.00 & 61.01 & 50.82 & 46.10 & 0.4074 \\*
L12 concat & 36.00 & 63.94 & 54.10 & 45.70 & 0.4223 \\*
L12 merge & 35.00 & 63.01 & 59.84 & 46.10 & 0.4094 \\
\addlinespace[2pt]
L16 mean & 45.50 & 65.00 & 64.75 & 52.63 & 0.3452 \\*
L16 concat & 47.00 & 68.00 & 62.30 & 53.36 & 0.3784 \\*
L16 merge & 44.50 & 62.01 & 63.11 & 52.55 & 0.3724 \\
\addlinespace[2pt]
L20 mean & 46.00 & 65.02 & 65.57 & 55.09 & 0.3820 \\*
L20 concat & 48.50 & 65.08 & 62.30 & 57.12 & 0.3851 \\*
L20 merge & 47.00 & 63.89 & 64.75 & 57.13 & 0.3725 \\
\addlinespace[2pt]
L24 mean & 48.00 & 64.02 & 64.75 & 56.62 & 0.3876 \\*
L24 concat & 49.50 & 68.07 & 66.39 & 56.84 & 0.3861 \\*
L24 merge & 46.50 & 64.95 & 63.11 & 57.85 & 0.3759 \\
\addlinespace[2pt]
L28 mean & 53.00 & 63.98 & 66.39 & 55.71 & 0.3960 \\*
L28 concat & 49.50 & 64.98 & 63.93 & 56.41 & 0.3721 \\*
L28 merge & 46.00 & 64.01 & 67.21 & 56.11 & 0.3923 \\
\addlinespace[2pt]
L32 mean & 51.00 & 59.98 & 62.30 & 56.74 & 0.3341 \\*
L32 concat & 51.00 & 60.05 & 61.48 & 55.39 & 0.3297 \\*
L32 merge & 50.00 & 63.06 & 63.93 & 56.40 & 0.3405 \\
\end{longtable}
\endgroup

\subsection{Head Activity, Redundancy, and Causal Interventions}
\label{sec:head-results}
We analyze the crop-trained L32 mean checkpoint without retraining.
At every decision node, the captured 16 per-head candidate-logit vectors are checked to average back to the original readout.
Three statistics answer different questions.
Centered RMS measures how strongly a head distinguishes candidates; its normalized share measures relative response magnitude; and removing the head measures its effect on predictions and quality.
For head $h$ and $K$ legal candidates,
\begin{equation}
 A_h=\sqrt{\frac1K\sum_i(s_{h,i}-\bar s_h)^2},\qquad
 z_{-h,i}=z_{\rm all,i}-\frac{s_{h,i}}{16}.
 \label{eq:head-intervention}
\end{equation}
Centering removes candidate-independent offsets.
The removal denominator stays 16, so the intervention does not amplify the remaining heads.
Each of 17 conditions---all heads plus 16 removals---follows its own numerical and crop trajectory.
Shared forward results are reused only when inputs and histories are identical.
Counting interventions reconstruct their own root-pruned maps from a captured full tree.

\begin{figure}[H]
\centering
\includegraphics[width=\textwidth]{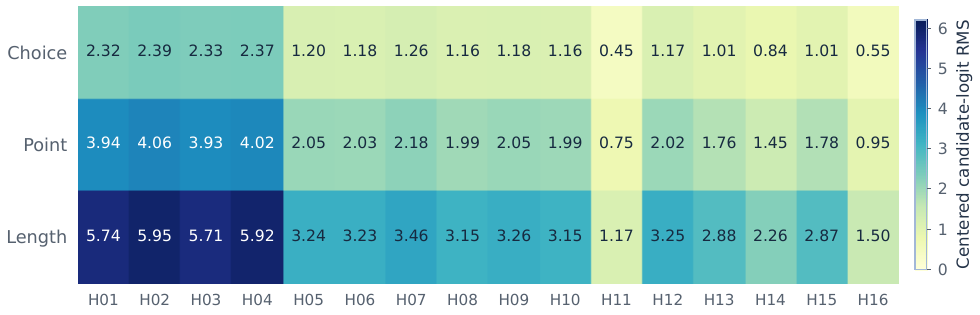}
\caption{\captionlead{Centered candidate-logit RMS for each head, averaged per QA along the all-head trajectory.}
Values share a common scale and measure response amplitude.
H02 is strongest and H11 weakest in all three task groups.}
\label{fig:heads}
\end{figure}

\begin{samepage}
Head amplitudes follow a similar ordering across tasks.
H02 has RMS 2.394, 4.065, and 5.954 for choice, point, and length; H11 has 0.447, 0.750, and 1.168, about 18--20\% of the strongest head.
Centered pairwise head-logit cosine averages are 0.994163, 0.999331, and 0.998055, with minimum pair averages 0.985885, 0.997355, and 0.991673.
The readout heads largely agree on candidate direction while differing in amplitude, which is more consistent with redundancy than clear task-specific head specialization.
\par
\end{samepage}

Single removals nevertheless change some complete paths.
The largest classification benchmark change is 0.20 points; Point-Bench can decrease by 0.5092 points; length MAE changes from $-0.003992$ to $+0.000279$\,m.
No head meets the predeclared near-zero screen: small score effect (at most 0.5 points per benchmark, or 1\% of length MAE), at most 1\% prediction changes in each task group, and at most 1\% response share.
The final model therefore retains all heads.
Single-head masking can change the logit scale used by thresholded counting maps; joint head pruning was not evaluated.

\paragraph{Complete intervention results.}
\label{app:heads}
\begin{table}[htbp]
\centering\small
\setlength{\tabcolsep}{5pt}
\caption{All 16 single-head interventions. Positive effect means retaining the head helps: full-minus-masked score, or masked-minus-full length MAE. Classification/pointing effects are benchmark-equal diagnostic means; RMS values average per question.}
\label{tab:head-interventions}
\begin{tabular}{lrrrrrr}
\toprule
\rowcolor{mindlabtablehead}
\mindlabtablehead{Head} & \mindlabtablehead{Class RMS} & \mindlabtablehead{Point RMS} & \mindlabtablehead{Length RMS} & \mindlabtablehead{Class $\Delta$ pp} & \mindlabtablehead{Point $\Delta$ pp} & \mindlabtablehead{Length $\Delta$ m} \\\midrule
H01 & 2.324 & 3.938 & 5.742 & -0.011 & +0.127 & +0.000000 \\
H02 & 2.394 & 4.065 & 5.954 & +0.009 & +0.127 & +0.000000 \\
H03 & 2.328 & 3.932 & 5.708 & +0.020 & +0.127 & +0.000000 \\
H04 & 2.368 & 4.022 & 5.917 & +0.009 & +0.127 & +0.000279 \\
H05 & 1.202 & 2.055 & 3.236 & +0.050 & +0.102 & -0.000346 \\
H06 & 1.180 & 2.032 & 3.232 & +0.002 & +0.102 & -0.003992 \\
H07 & 1.264 & 2.176 & 3.463 & +0.002 & +0.127 & -0.000346 \\
H08 & 1.156 & 1.986 & 3.154 & +0.030 & +0.102 & -0.003646 \\
H09 & 1.183 & 2.050 & 3.258 & +0.009 & +0.076 & -0.000067 \\
H10 & 1.159 & 1.989 & 3.154 & +0.012 & +0.102 & -0.003646 \\
H11 & 0.447 & 0.750 & 1.168 & +0.051 & +0.000 & -0.003646 \\
H12 & 1.167 & 2.016 & 3.250 & -0.011 & +0.102 & -0.000346 \\
H13 & 1.010 & 1.765 & 2.883 & -0.039 & +0.102 & +0.000182 \\
H14 & 0.841 & 1.448 & 2.258 & -0.018 & +0.102 & -0.003809 \\
H15 & 1.011 & 1.779 & 2.873 & -0.034 & +0.102 & -0.003992 \\
H16 & 0.551 & 0.949 & 1.504 & +0.002 & +0.051 & -0.000346 \\
\bottomrule
\end{tabular}
\end{table}

For ordinary multi-stage QA, node statistics are averaged across the all-head predicted path.
Counting nodes use parent probability mass weights over the visited tree, normalized per question; questions then receive equal weight.
The absolute RMS is computed before this aggregation.
Relative response shares are normalized within each node, so normalizing the final mean RMS would produce a different statistic.

Task-summary score effects average benchmark scores equally, whereas RMS, response shares, and prediction-change rates average individual QA.
\FloatBarrier

\section{Extended Spatial Refinement Studies}
\label{app:visual-studies}
\subsection{Visual Refinement: New Pixels, Replayed Features, or Direct Tokens?}
\label{app:visual-comparisons}
Hierarchical point classification must preserve enough visual detail to distinguish increasingly small regions.
We therefore separate three questions: whether crop-trained weights can operate without crops, whether matching training to a cheaper visual input recovers quality, and whether spatial candidates should be text regions or visual tokens.
All learned variants start from the same official Qwen base.
Training QA, updates, optimizer, and per-QA loss normalization are held fixed; visual computation is an intended variable, so the runs do not have equal FLOPs.

\paragraph{Removing crops from fixed weights.}
The inference-only intervention keeps three nine-way textual decisions but removes both appended pixel crops.
RefSpatial drops from 51.00 to 9.00, Where2Place from 59.98 to 32.00, and RoboSpatial pointing from 62.30 to 20.49 (\cref{tab:nocrop}).
Adding an original-image hybrid prefix cache leaves these aggregate scores unchanged.
The immediate loss is therefore not repaired by reusing the original context more efficiently.
However, this comparison changes both visual detail and the input distribution seen during training; it cannot determine the potential of a separately trained no-crop model.

\begin{table}[htbp]
\centering\small
\caption{Fixed crop-trained native-mean checkpoint: removing new pixel crops, with and without a reusable original-image prefix.
All three textual grid rounds remain.
The crop control was not timed in this run.}
\label{tab:nocrop}
\begin{tabular}{lrrr}
\toprule
\rowcolor{mindlabtablehead}
\mindlabtablehead{Metric} & \mindlabtablehead{Crop} & \mindlabtablehead{Original only} & \mindlabtablehead{Original + cache} \\\midrule
RefSpatial (\%) & 51.00 & 9.00 & 9.00\\
Where2Place (\%) & 59.98 & 32.00 & 32.00\\
RoboSpatial pointing (\%) & 62.30 & 20.49 & 20.49\\
Point-Bench macro (\%) & 56.84 & 21.96 & 21.96\\
Counting success (\%) & 39.29 & 4.08 & 4.08\\
Count MAE & 1.7296 & 4.0459 & 4.0561\\
Single-point time (s/question) & --- & 0.1519 & 0.2910\\
Counting time (s/question) & --- & 1.8144 & 1.5608\\
\bottomrule
\end{tabular}
\end{table}

\paragraph{Matched training with original-image context.}
The original-image variant removes crop placeholders during training as well as inference, while retaining the selected rectangle and region options in text.
Training still has three supervised decisions per pointing QA, and choice/scalar token sequences and targets remain unchanged.
RefSpatial recovers to 46.50 and Point-Bench macro to 52.55 (\cref{tab:roi}).
This recovery shows that much of the inference-only collapse is an input-distribution mismatch.
The remaining gap to crop re-encoding suggests that original-image context and region text do not fully replace newly resolved local visual information in this setting.

\paragraph{ROI feature replay.}
The next variant encodes the original pixels once per forward and appends existing merged visual tokens whose spatial supports have positive overlap with the selected ROI.
Tokens retain row-major source order and their original multimodal rotary positions; there is no interpolation to a new fixed grid.
This exposes region-focused representations without new pixel encoding.
Boundary tokens may still include content outside the ROI, and copying cannot recover detail absent from the first visual tokenization.

\paragraph{ROI attention K/V replay.}
A second replay variant appends the same visual tokens but replaces their keys and values in complete full-attention layers L4--L28 with differentiably gathered K/V from original-image positions after rotary encoding.
Queries, gates, residuals, and FFNs execute normally.
Source K/V remain in the current computation graph, and the original-image prefix precedes all question histories.

The matched comparison is reported in \cref{tab:roi}.
Neither replay scheme closes the overall gap to crop re-encoding.
ViT replay reaches 47.00 RefSpatial and 52.87 Point-Bench, only slightly above matched original-image training.
KV replay improves Where2Place to 68.95 but decreases RoboSpatial pointing to 54.92 and increases length MAE to 0.4097\,m.
Choice and numerical scores also change despite identical non-point inputs, consistent with interactions through the jointly trained backbone.
These outcomes support crop re-encoding as the final common representation.

\paragraph{Direct visual-token candidates.}
We train two further models, KEV and native mean, whose point candidates are the original image's merged visual tokens in the \emph{language sequence}.
The query is the question-end decision state; keys use the same language-side projections as their respective readout.
Let $H'=H/s$ and $W'=W/s$ denote the dimensions of the merged visual-token grid; each cell is a candidate.
Cell $(r,c)$ decodes to $((c+1/2)/W', (r+1/2)/H')$ in original normalized coordinates.

Point training uses one decision.
Ground-truth points are mapped to occupied cells by floor-and-clamp, with equal target mass across distinct occupied cells.
Repeated points in one cell do not multiply its weight, and target masks do not enlarge the positive set.
This preserves one unit of loss per QA while changing both the spatial representation and number of stages.
The one-pass method removes crop encoding, but its grid resolution depends on image aspect ratio and token budget; it cannot resolve two instances that share the same visual cell.

\begin{table}[htbp]
\centering\small
\setlength{\tabcolsep}{5pt}
\caption{Direct visual-token pointing with KEV (K) or native mean (N). Each model shares weights/logits across argmax, components, and task-based decoding. Task routing was specified after the first decoding experiment; Point-Bench is macro.}
\label{tab:visual-decoders}
\begin{tabular}{lrrrrrr}
\toprule
\rowcolor{mindlabtablehead}
\mindlabtablehead{Metric} & \mindlabtablehead{K argmax} & \mindlabtablehead{K comp.} & \mindlabtablehead{K task} & \mindlabtablehead{N argmax} & \mindlabtablehead{N comp.} & \mindlabtablehead{N task} \\\midrule
RefSpatial & 38.50 & 35.14 & 38.50 & 45.50 & 43.65 & 45.50 \\
Where2Place & 67.96 & 64.43 & 64.43 & 60.91 & 58.99 & 58.99 \\
RoboSpatial pointing & 56.56 & 57.38 & 57.38 & 60.66 & 63.11 & 63.11 \\
Point-Bench (macro) & 46.46 & 26.81 & 49.12 & 50.76 & 32.71 & 54.63 \\
PB counting & 13.27 & 26.53 & 26.53 & 12.76 & 32.14 & 32.14 \\
\bottomrule
\end{tabular}
\end{table}

\begin{table}[htbp]
\centering\small
\setlength{\tabcolsep}{5pt}
\caption{Visual-token pointing versus hierarchical crop pointing. Visual-token results use task-based single/multiple-point routing established after the initial decoding experiment. This routing differs from the grid baseline on Where2Place and RoboSpatial; it is an exploratory comparison.}
\label{tab:visual-tokens}
\begin{tabular}{lrrrr}
\toprule
\rowcolor{mindlabtablehead}
\mindlabtablehead{Metric} & \mindlabtablehead{KEV grid} & \mindlabtablehead{KEV visual} & \mindlabtablehead{Mean grid} & \mindlabtablehead{Mean visual} \\\midrule
SAT real & 76.33 & 76.67 & 78.00 & 77.00 \\
CV-Bench & 85.31 & 83.95 & 85.38 & 84.52 \\
RoboSpatial VQA & 68.54 & 64.32 & 68.06 & 62.50 \\
BLINK val & 62.99 & 63.87 & 63.04 & 62.76 \\
MMSI-Bench & 25.10 & 25.50 & 27.70 & 30.20 \\
RefSpatial & 46.00 & 38.50 & 51.00 & 45.50 \\
Where2Place & 61.02 & 64.43 & 59.98 & 58.99 \\
RoboSpatial pointing & 63.11 & 57.38 & 62.30 & 63.11 \\
Point-Bench (macro) & 57.33 & 49.12 & 56.74 & 54.63 \\
VST-300 MAE (m) & 0.3614 & 0.3350 & 0.3341 & 0.3614 \\
\bottomrule
\end{tabular}
\end{table}

We decode each saved visual-logit grid with global argmax, global connected components, and a task-based rule.
The latter uses argmax for RefSpatial and non-counting Point-Bench, and components for Where2Place, RoboSpatial, and counting.
The rule was specified after the first decoder comparison; all three columns for a model share weights and logits, with no extra training or forward passes for the task-based recombination.

The decoder ablation explains why a universal multi-point policy is unsuitable (\cref{tab:visual-decoders}).
For native visual tokens, components improve counting from 12.76 to 32.14 but reduce overall Point-Bench macro from 50.76 to 32.71: extra peaks harm tasks that expect one point.
Task-based routing restores the single-point categories and reaches 54.63.
It still trails hierarchical mean on RefSpatial (45.50 versus 51.00) and Point-Bench (54.63 versus 56.74).
Cross-task effects also differ between KEV and native readouts (\cref{tab:visual-tokens}); for example, native visual-token training improves MMSI but worsens length MAE.
The experiments favor retaining hierarchical crops for the final model.

\subsection{Multi-Point Decoding: Probability Trees versus Stop Decisions}
\label{sec:tree-results}
The first counting extension repeatedly asks a continue/stop choice and, when continuing, invokes single-point prediction with the already selected coordinates in the prompt.
This composes existing interfaces without training on exhaustive point-set completion.
It relies on coverage and stopping behavior from a model trained primarily on single targets.
On all 196 Point-Bench counting questions, it reaches 19.39\% official success and count MAE 11.64.
A small set of long overproduced sequences inflates the mean predicted count to 10.48 even though many other examples undercount.

\begin{table}[htbp]
\centering\small
\setlength{\tabcolsep}{5pt}
\caption{Point-Bench counting: all 196 questions, fixed L32 mean weights, no retraining. Full/pruned are an actual paired run; historical continue/stop quality uses a different execution setting and has no comparable timing here.}
\label{tab:tree}
\begin{tabular}{lrrr}
\toprule
\rowcolor{mindlabtablehead}
\mindlabtablehead{Metric} & \mindlabtablehead{Continue/stop} & \mindlabtablehead{Full tree} & \mindlabtablehead{Root-pruned} \\\midrule
Official success (\%) & 19.39 & 38.78 & 38.78 \\
Exact count (\%) & 22.45 & 46.43 & 46.43 \\
Count MAE & 11.643 & 1.735 & 1.740 \\
Predicted points & 10.485 & 3.342 & 3.337 \\
Decision nodes & -- & 91.00 & 51.10 \\
Evaluated cells & -- & 729.00 & 405.83 \\
Omitted mass (\%) & -- & 0.000 & 0.588 \\
Seconds / question & -- & 4.716 & 2.638 \\
\bottomrule
\end{tabular}
\end{table}

The probability-tree alternative in \cref{sec:pointing} measures the spatial mass of all potential paths and extracts components without a learned stop query.
Both full and root-pruned trees reach 38.78\% success and 46.43\% exact-count accuracy in the dedicated paired experiment (\cref{tab:tree}).
Root pruning reduces mean node count from 91 to 51.10, a 43.84\% reduction, while omitting only 0.588\% probability mass on average.
Measured per-question time decreases from 4.716 to 2.638 seconds, a 1.79$\times$ ratio under this execution setup.

The two modes are separately executed, alternate order by question index, and use untimed warmup, batches of 16 nodes, eight GPUs, visual deduplication, and no shared language prefix.
Timings include processing and GPU computation but exclude loading, warmup, and saving.
Predicted counts agree on 195/196 questions and success on all 196.
Shadow pruning of the saved full-tree logits changes no counts; the one count discrepancy in the actual paired runs is therefore associated with changed numerical execution rather than omitting a root region alone.
The later consolidated benchmark rerun has slightly different full-tree scores and timings; its own values are retained in the overall matrix instead of replacing this paired control.

\begin{figure}[tbp]
\centering
\includegraphics[width=\textwidth]{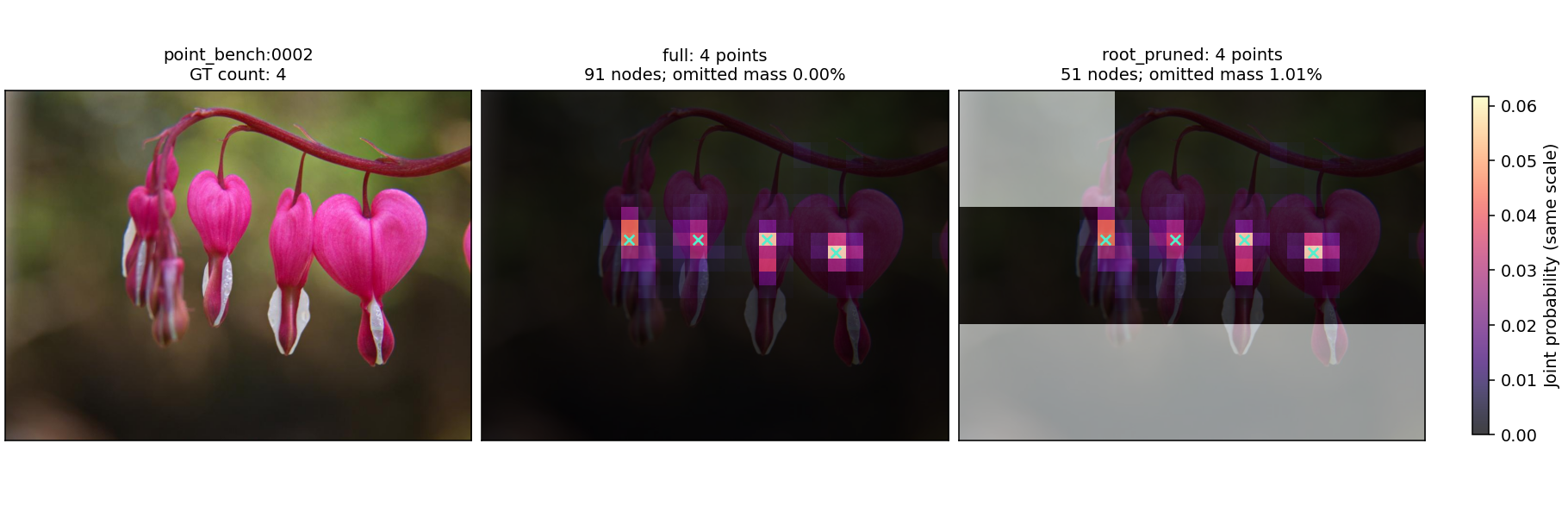}
\caption{\captionlead{Probability-tree example from the first request in the preselected 12-case gallery.}
Full expansion uses 91 nodes; root pruning uses 51 and marks unvisited regions in gray.
Both return four points in this case.
Example from Point-Bench~\citep{pointarena}.}
\label{fig:tree-example}
\end{figure}

\Cref{tab:tree-sensitivity} reports fixed 5\%, 10\%, and 20\% peak-threshold sensitivity.
Success is 30.10, 38.78, and 37.24\%, respectively, for both modes in this record.
The original 10\% threshold remains the primary rule; the sensitivity table demonstrates that point-set quality depends on map postprocessing as well as candidate accuracy.
The author scorer requires exact count and every predicted point inside the union mask, which does not by itself certify one prediction per object instance.

\begin{table}[htbp]
\centering\small
\setlength{\tabcolsep}{5pt}
\caption{Offline component-threshold sensitivity on the retained maps. Success and exact-count columns are percentages. The primary threshold is fixed at 10\%.}
\label{tab:tree-sensitivity}
\begin{tabular}{lrrrrrr}
\toprule
\rowcolor{mindlabtablehead}
\mindlabtablehead{Peak threshold} & \mindlabtablehead{Full success} & \mindlabtablehead{Pruned success} & \mindlabtablehead{Full count} & \mindlabtablehead{Pruned count} & \mindlabtablehead{Full MAE} & \mindlabtablehead{Pruned MAE} \\\midrule
5\% & 30.10 & 30.10 & 34.69 & 34.69 & 2.041 & 2.041 \\
10\% & 38.78 & 38.78 & 46.43 & 46.43 & 1.735 & 1.740 \\
20\% & 37.24 & 37.24 & 43.88 & 43.88 & 1.765 & 1.760 \\
\bottomrule
\end{tabular}
\end{table}

\section{Extended Numerical Studies}
\label{app:numerical-studies}
\subsection{Numerical Design: Coverage, Granularity, and Decoding}
\label{app:length-results}
Numerical discretization has two sources of error: representing a true value with a finite cell center, and predicting the correct cell from the image.
The following experiments measure these separately.
Codebook candidates are designed from training labels, and the 4/8 codebook below is used for the final model.
All learned comparisons, however, retain the old 72K-QA gradient-training budget, including only 22K length examples.

\paragraph{Coverage comes before precision.}
The historical codebook has 12 positive intervals ending at 200\,m, two special roots, and 32 fine cells per positive interval.
It covers all old 72K length labels but overflows on 12K expanded-corpus labels.
The scale audit compares uniform, log, quantile, task-specific, and hierarchical allocations on all 1.6M accepted lengths (\cref{tab:length-oracles}).
These are label encode/decode oracles, assuming every classification decision is correct.

Uniform 960-cell spacing up to 16,384\,m gives MAE 6.4545\,m because most labels are short.
Quantile spacing improves the proportion within 1\,cm to 80.78\%, yet produces MAE 4.2371\,m and maximum error 3935.55\,m: a dense central distribution leaves the sparse tail poorly represented.
Shared log allocations balance these scales more effectively.
The 30-by-32 two-stage log1p proposal reaches 0.0391\,m oracle MAE and 21.39\,m maximum error, compared with 1.6548 and 8320.10\,m for the historical codebook.
Task-specific log allocation offers a small average improvement but introduces question-type routing and poorly supported rare categories (e.g., only five diameter and four thickness labels).
We therefore keep a shared length codebook and the existing linear within-interval decoder.

\paragraph{Testing the 30-by-32 proposal as a model.}
A new L32 mean model uses 30 positive log1p intervals and 32 fine cells per interval (960 leaves), preserving all other 72K settings and the exact non-numerical stage texts and mappings.
On those training labels, oracle MAE falls from 0.006591 to 0.003891\,m and P99 error from 0.045625 to 0.023542\,m, but worst-case error rises from 0.7475 to 0.9800\,m.
The new numerical candidate text adds 527.43 tokens on average; the longest checked scalar sequence is 2,068 tokens, within the 4,096-token budget.
Coverage and average quantization improve at a real input-cost increase.

The learned model reduces VST-300 MAE from 0.335420 to 0.327732\,m, only 2.29\%, and lowers all five classification benchmark scores.
The much larger oracle gain does not translate directly into prediction accuracy.
This motivates reducing fine-stage difficulty rather than adding still more representational resolution.

\paragraph{Reducing the fine-stage candidate count.}
The 4/8 design in \cref{eq:tolerance} keeps the same coarse intervals and range but reduces positive leaves from 960 to 216.
In the full length census, 108K positive labels use four fine candidates, 1.5M use eight, and 382 terminate at zero.
Both the 32-fine and 4/8 representations meet the specified $\max(1\,\mathrm{cm},5\%\,y)$ oracle tolerance on every audited label, with no overflow.
Coarser cells increase oracle MAE from 0.039102 to 0.154041\,m and maximum error from 21.39 to 59.55\,m.
The 4/8 codebook exchanges quantization resolution for a smaller candidate set.

A second independently initialized model changes only these fine cardinalities.
Choice, point, count, and angle implementations remain fixed; no length batching optimization is introduced in this quality comparison.
The estimated scalar sequence length is bounded by 1,514 tokens.
Training takes 1,369.33 seconds versus 1,604.98 for the prior 30-by-32 run under their recorded timing scopes.
The shorter candidate representation is consistent with lower work, although these single-run training times are not a dedicated timing ablation.

The learned length errors are reported in \cref{tab:length-diagnostics}.
\begin{table}[H]
\centering\small
\setlength{\tabcolsep}{5pt}
\caption{Fixed internal length diagnostic. 4/8 argmax is the decoder used by \method{}; 4/8 exp.\ decodes the same logits with a fine-stage conditional expectation. Coarse decoding is argmax in every condition.}
\label{tab:length-diagnostics}
\begin{tabular}{lrrrr}
\toprule
\rowcolor{mindlabtablehead}
\mindlabtablehead{VST-300 diagnostic} & \mindlabtablehead{Historical} & \mindlabtablehead{$30\times32$} & \mindlabtablehead{4/8 argmax} & \mindlabtablehead{4/8 exp.} \\\midrule
MAE (m), lower & 0.335420 & 0.327732 & 0.320207 & 0.329955 \\
AbsRel (\%), lower & 24.64 & 31.04 & 29.03 & 30.04 \\
$|e|\leq 0.1$ m (\%), higher & 56.67 & 58.33 & 59.00 & 55.67 \\
$|e|/y\leq 25\%$ (\%), higher & 69.46 & 69.46 & 68.12 & 70.81 \\
\bottomrule
\end{tabular}
\end{table}

\begin{table}[htbp]
\centering\small
\setlength{\tabcolsep}{5pt}
\caption{Length-codebook ablation, independently trained on the same 72K QA. The historical checkpoint is re-evaluated within this experiment. All point and choice decoders are unchanged.}
\label{tab:length-quality}
\begin{tabular}{lrrr}
\toprule
\rowcolor{mindlabtablehead}
\mindlabtablehead{Metric} & \mindlabtablehead{Historical} & \mindlabtablehead{$30\times32$} & \mindlabtablehead{$30\times(4/8)$} \\\midrule
SAT real & 78.00 & 72.33 & 80.67 \\
CV-Bench & 85.39 & 85.08 & 85.21 \\
RoboSpatial VQA & 68.06 & 64.87 & 66.30 \\
BLINK val & 63.10 & 60.94 & 62.52 \\
MMSI-Bench & 27.70 & 25.40 & 27.60 \\
RefSpatial & 51.00 & 49.50 & 52.00 \\
Where2Place & 59.98 & 61.96 & 65.02 \\
RoboSpatial pointing & 62.30 & 60.66 & 63.93 \\
Point-Bench (macro) & 56.74 & 56.83 & 57.14 \\
VST-300 MAE (m) & 0.3354 & 0.3277 & 0.3202 \\
\bottomrule
\end{tabular}
\end{table}

The 4/8 argmax model achieves MAE 0.320207\,m, 4.54\% below the historical control and 2.30\% below 30-by-32.
Errors within 10\,cm increase to 59.00\%.
All nine public primary scores improve relative to 30-by-32; relative to the historical codebook, SAT and the four pointing scores improve, while four other classification scores remain slightly lower.
The historical codebook still has better mean relative error (24.64\% versus 29.03\%), emphasizing that MAE and relative accuracy favor different parts of the value distribution.

\paragraph{Does averaging fine-cell predictions help?}
Using exactly the same 4/8 checkpoint and saved logits, we compare fine argmax with a temperature-1 conditional expectation,
\begin{equation}
 \hat y_{\rm exp}=\sum_{k=0}^{F_{\hat j}-1}
       p(k\mid\hat j,I,q)\left[e_{\hat j}+
          \frac{k+1/2}{F_{\hat j}}(e_{\hat j+1}-e_{\hat j})\right].
\end{equation}
The root remains argmax; zero and overflow behavior are unchanged.
The expectation decoder raises MAE to 0.329955\,m (+3.04\%) and reduces 10\,cm accuracy to 55.67\%, despite improving the 25\%-relative-error hit rate to 70.81\%.
\method{} therefore decodes lengths with argmax at both stages; the other reported length results and the final model use this decoding rule.
All non-length outputs are identical in this same-logit comparison.

\paragraph{Complete reconstruction oracles.}
\label{app:numerical}
\begin{table}[htbp]
\centering\small
\setlength{\tabcolsep}{5pt}
\caption{Training-label reconstruction on all 1.6M lengths, assuming the correct discrete cell is selected. The 960-leaf alternatives differ in allocation; only the hierarchical log1p proposal is tested in the subsequent matched model experiment. Task-specific routing is an unadopted proposal.}
\label{tab:length-oracles}
\begin{tabular}{lrrrrr}
\toprule
\rowcolor{mindlabtablehead}
\mindlabtablehead{Representation} & \mindlabtablehead{MAE (m)} & \mindlabtablehead{P99 (m)} & \mindlabtablehead{Max (m)} & \mindlabtablehead{$\leq1$ cm (\%)} & \mindlabtablehead{$\leq1\%$ rel.} \\\midrule
Historical 200 m & 1.6548 & 1.2384 & 8320.10 & 59.33 & 80.29 \\
Uniform 960 & 6.4545 & 8.5012 & 8.53 & 0.06 & 1.12 \\
Log 384 & 0.0939 & 1.5758 & 43.10 & 47.53 & 52.36 \\
Log 960 & 0.0379 & 0.6371 & 47.14 & 69.86 & 99.92 \\
Quantile 960 & 4.2371 & 3.0380 & 3935.55 & 80.78 & 95.50 \\
Hierarchical log1p 960 & 0.0391 & 0.6657 & 21.39 & 69.42 & 99.84 \\
Task-specific log 960 & 0.0372 & 0.6308 & 47.14 & 70.93 & 99.99 \\
\bottomrule
\end{tabular}
\end{table}

The representation comparison includes a 384-cell shared logarithmic control matching the historical positive-cell budget and several 960-cell allocations.
The selected hierarchical proposal differs from a globally logarithmic leaf grid: its roots are log1p-spaced, but each root's fine cells are linear.
The actual final 216-cell variant is documented in \cref{sec:numeric,sec:length-results}.
All oracle labels come from the expanded training corpus, not from evaluation targets.

\FloatBarrier

\section{Training Efficiency Diagnostics}
\label{app:training-measurements}
\subsection{Length Grouping Separates Padding from Rank Imbalance}
The final scheduling mechanism is defined in \cref{app:training-execution}.
Its performance study holds global batch 256, microbatch 8, four accumulation steps, eight L20D GPUs, the old 72K data/codebook, and the timing instrumentation fixed.
Four fresh runs each process 100 updates and 25,600 QA: original order, local length grouping, global grouping with rank balancing, and global balancing with a 128-QA buffer.
The last 92 updates contain the same 23,552 measured QA; the first eight are excluded for warmup.
The buffer is drained at both measurement boundaries, and slow steps remain in the timing record.

The controlled scheduling results are reported in \cref{tab:efficiency}.
\begin{figure}[H]
\centering
\includegraphics[width=.95\textwidth]{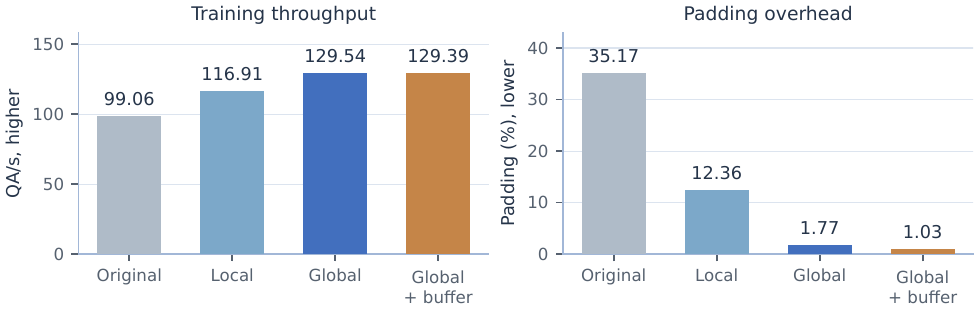}
\caption{\captionlead{Controlled scheduling comparison on the historical 72K corpus.}
Local grouping reduces padding but leaves more rank imbalance; global grouping and balancing deliver the largest throughput gain.
A cross-update buffer reduces padding slightly further without increasing throughput.}
\label{fig:efficiency}
\end{figure}

Local grouping increases throughput from 99.06 to 116.91 QA/s and reduces padding from 35.17\% to 12.36\%.
However, mean gradient-ready spread between ranks increases from 115.0 to 306.6\,ms, and tail communication/wait time from 39.5 to 157.6\,ms.
Minimizing padding independently on each rank can therefore leave unequal workloads at synchronization.
Global grouping plus balancing reduces padding to 1.77\%, reduces ready-time spread relative to local grouping to 145.7\,ms, and reaches 129.54 QA/s: a 30.78\% throughput increase or 23.53\% less measured time for equal QA volume.
Ready-time spread and tail time remain above the original-order values, so this gain reflects the combined scheduling changes rather than an isolated reduction in synchronization waiting.

The 128-QA buffer lowers padding to 1.03\% but achieves 129.39 QA/s.
Its step-time coefficient of variation increases from 6.07\% to 19.77\%; even after aggregating pairs of updates into equal 512-QA units, variation is 6.30\% versus 4.61\% without the buffer.
This identifies diminishing returns from further packing and supports the unbuffered final layout.
Tail measurements include communication and waiting; no downstream accuracy evaluation is performed in these short runs.

\subsection{Microbatch Feasibility and Global Batch Size}
\begin{table}[htbp]
\centering\small
\setlength{\tabcolsep}{5pt}
\caption{Expanded-data memory feasibility. The original six cases use 4,096 warmup and 8,192 measured QA; the additional global384/micro12 trial also fails. Partial timings from failed configurations are excluded. The successful rows are resummarized on a common window in Table~\ref{tab:batch}.}
\label{tab:microbatch-matrix}
\begin{tabular}{lrrrrrr}
\toprule
\rowcolor{mindlabtablehead}
\mindlabtablehead{Global} & \mindlabtablehead{Micro} & \mindlabtablehead{Accum.} & \mindlabtablehead{Outcome} & \mindlabtablehead{s/update} & \mindlabtablehead{QA/s} & \mindlabtablehead{Peak GB} \\\midrule
256 & 8 & 4 & Complete & 2.3007 & 111.27 & 242.03 \\
512 & 8 & 8 & Complete & 3.7629 & 136.07 & 242.05 \\
256 & 16 & 2 & OOM & -- & -- & -- \\
512 & 16 & 4 & OOM & -- & -- & -- \\
256 & 32 & 1 & OOM & -- & -- & -- \\
512 & 32 & 2 & OOM & -- & -- & -- \\
384 & 12 & 4 & OOM & -- & -- & -- \\
\bottomrule
\end{tabular}
\end{table}

The expanded-data profile uses a fixed 12,800-QA prefix-stratified pool in the then-proposed 40/25/25/5/5 task mixture, the new 4/8 length codebook, and fresh base initialization.
All conditions use global length grouping, rank balancing, no buffer, no activation checkpointing~\citep{gradcheckpoint}, and the same precision and optimizer.
The original six-case matrix crosses global256/512 with micro8/16/32.
Its measurement window follows 4,096 warmup QA and includes 8,192 common timed QA.
Only micro8 succeeds: 111.27 QA/s at global256 and 136.07 at global512.
Micro16 and micro32 fail in both cases.
An additional global384/micro12 trial also fails.

\begin{table}[htbp]
\centering\small
\setlength{\tabcolsep}{5pt}
\caption{New-data profiling with the 4/8 length codebook: a common 7,680-QA window after 4,608 QA warmup. The 12,800-QA pool is a stratified prefix sample, not the natural full-data distribution. Peak is the maximum per-rank allocated GPU memory, in decimal GB. Microbatch 16 and 32 failed with OOM in both original 256/512 conditions.}
\label{tab:batch}
\begin{tabular}{lrrrrrr}
\toprule
\rowcolor{mindlabtablehead}
\mindlabtablehead{Global} & \mindlabtablehead{Micro} & \mindlabtablehead{Accum.} & \mindlabtablehead{Timed steps} & \mindlabtablehead{s/update} & \mindlabtablehead{QA/s} & \mindlabtablehead{Peak GB} \\\midrule
256 & 8 & 4 & 30 & 2.3265 & 110.04 & 242.03 \\
512 & 8 & 8 & 15 & 3.7791 & 135.48 & 242.05 \\
768 & 8 & 12 & 10 & 5.4944 & 139.78 & 242.03 \\
\bottomrule
\end{tabular}
\end{table}

The global256/512/768 profiles use a common window of 4,608 warmup QA followed by 7,680 timed QA.
The window has 30, 15, and 10 updates for the three global batches, respectively.
\Cref{tab:batch} reports 110.04, 135.48, and 139.78 QA/s for these batch sizes.
Peak allocated memory is approximately 242 decimal GB per rank for the successful configurations.

The experiment shows that additional accumulation can raise the global batch without increasing the forward microbatch beyond its memory limit.
Global768 improves throughput over global512 by 3.17\% over ten measured updates.
Moreover, all 16 executed global768 updates fall within the zero-backbone-LR period of its short schedule.
Backward computation and Adam moment updates still execute during these steps.
The final recipe therefore selects global512/micro8/accumulation8.

\end{document}